# How Robust is GPT-3.5 to Predecessors? A Comprehensive Study on Language Understanding Tasks

**Xuanting Chen**[★,*], **Junjie Ye**[★,*], **Can Zu**[★], **Nuo Xu**[★], **Rui Zheng**[★],
**Minlong Peng**[★], **Jie Zhou**[★], **Tao Gui**[♦,†], **Qi Zhang**[★,♣], **Xuanjing Huang**[★]

[★] School of Computer Science, Fudan University, Shanghai, China
[♦] Institute of Modern Languages and Linguistics, Fudan University, Shanghai, China
[♣] Shanghai Collaborative Innovation Center of Intelligent Visual Computing, Fudan University
xuantingchen21@m.fudan.edu.cn
{jjye19,tgui,qz,xjhuang}@fudan.edu.cn

## Abstract

The GPT-3.5 models have demonstrated impressive performance in various Natural Language Processing (NLP) tasks, showcasing their strong understanding and reasoning capabilities. However, their robustness and abilities to handle various complexities of the open world have yet to be explored, which is especially crucial in assessing the stability of models and is a key aspect of trustworthy AI. In this study, we perform a comprehensive experimental analysis of GPT-3.5, exploring its robustness using 21 datasets (about 116K test samples) with 66 text transformations from TextFlint that cover 9 popular Natural Language Understanding (NLU) tasks. Our findings indicate that while GPT-3.5 outperforms existing fine-tuned models on some tasks, it still encounters significant robustness degradation, such as its average performance dropping by up to 35.74% and 43.59% in natural language inference and sentiment analysis tasks, respectively. We also show that GPT-3.5 faces some specific robustness challenges, including robustness instability, prompt sensitivity, and number sensitivity. These insights are valuable for understanding its limitations and guiding future research in addressing these challenges to enhance GPT-3.5's overall performance and generalization abilities.

## 1 Introduction

Large Language Models (LLMs), such as FLAN (Wei et al., 2022), GPT-3 (Brown et al., 2020), OPT (Zhang et al., 2022), and PaLM (Chowdhery et al., 2022), have exhibited remarkable capabilities in approaching human-level performance in terms of dialogue quality, understanding, and reasoning. They can perform various tasks by inputting appropriate prompts or instructions, without any parameter modifications (Kaplan et al., 2020; Thoppilan et al., 2022). These advancements are driving the field of NLP toward the paradigm of Artificial General Intelligence (AGI) (Goertzel, 2014; Fei et al., 2022). Among all existing LLMs, the Generative Pre-trained Transformer (GPT) (Brown et al., 2020) series models are particularly popular because of their exceptional abilities.

The remarkable achievements of LLMs have spurred extensive investigations into their capabilities. Previous research has focused on examining LLMs' performance on specific capabilities, including commonsense and logical reasoning (Qin et al., 2023), multilingualism and multimodality (Bang et al., 2023), theory of mind (Kosinski, 2023), and mathematical (Frieder et al., 2023) abilities. However, despite demonstrating high scores on the hold-out test sets and showing significant abilities in these aspects, it remains unclear whether LLMs are robust enough to handle the complexities of the open world. Evaluating robustness is critical for ensuring the model's reliability in the real world, particularly in security-related scenarios, such as autopilot and medical diagnosis. However, a systematic evaluation of robustness has yet not been well-studied on LLMs.

Approaches to textual robustness evaluation measure the models' stability in the presence of slight modifications to the input, that occurs at the character, word, or sentence level. One line of work has focused on adversarial attacks to generate semantically equivalent adversaries (Jin et al., 2020; Ebrahimi et al., 2018). However, these methods can be computationally expensive and time-consuming, and may not be feasible for a large-scale evaluation. Therefore, another line of work has designed general robustness evaluation benchmarks that involve manually designing or modifying data, including CheckList (Ribeiro et al., 2020), Dynabench (Potts et al., 2020), and TextFlint (Gui et al., 2021). In this paper, we chose TextFlint, a unified toolkit that encompasses numerous NLP tasks and linguistically based text transformations, to conduct a comprehensive evaluation of robustness. In addition, we select the most advanced GPT-3.5 model, *text-davinci-003*, as our subject of analysis. Its remarkable performance and ease of use from the OpenAI API enable us to discover the robustness limitations of LLMs in a large-scale evaluation and provide valuable insights for future research.

---

[*] Equal contribution.
[†] Corresponding authors.

In this research, We empirically evaluate GPT-3.5 on 21 datasets (overall 115715 test samples) covering 9 popular NLU tasks and including 66 text transformations from TextFlint. Our investigation focuses on the following aspects: (1) GPT-3.5's robustness and performance in zero-shot and few-shot scenarios; (2) The exploration of specific robustness challenges and limitations of GPT-3.5; (3) Comparisons of GPT-3.5 to finetuned models; (4) The differences between the GPT-3 (*text-davinci-001*) and GPT-3.5 series (*text-davinci-002*, *text-davinci-003*) models.

Our study has yielded the following findings:

- **Competitive results on test sets:** GPT-3.5 achieves state-of-the-art results in some NLU tasks compared to supervised models fine-tuned with task-specific data. In particular, GPT-3.5 performs well in reading comprehension and sentiment analysis tasks, but faces significant challenges in sequence tagging and relation extraction tasks (Figure 1).

- **Lack of robustness:** GPT-3.5 still encounters significant robustness degradation, such as its average performance dropping by up to 35.74% and 43.59% in natural language inference (Section 4.3.1) and sentiment analysis (Section 4.4) tasks, respectively. However, it is worth noting that GPT-3.5 achieves remarkable robustness on certain tasks, such as reading comprehension (Section 4.1), and WSC (Section 4.6) tasks.

- **Robustness instability:** In few-shot scenarios, GPT-3.5's robustness improvement varies greatly across different tasks. For example, GPT-3.5 shows significant improvement in aspect-based sentiment analysis tasks while the robustness actually decreases in natural language inference (Section 4.3.1) and semantic matching (Section 4.3.2) tasks.

- **Prompt sensitivity:** Changes in input prompts[1] have a significant impact on the results, and GPT-3.5's robustness to prompt variations still requires improvement. In sentiment analysis (Section 4.4), natural language inference (Section 4.3.1), semantic matching (Section 4.3.2) , and reading comprehension (Section 4.1) tasks, GPT-3.5 exhibits high variance across different prompts.

- **Number sensitivity:** GPT-3.5 is more sensitive to numerical inputs than pre-training fine-tuning models. For example, in the *NumWord* transformation, which involves replacing numerical words in sentences with different numerical values, GPT-3.5 exhibits a significantly high level of sensitivity (Section 4.3).

- **Task labels sensitivity:** We speculate that the task construction during the instruction tuning stage may significantly impact the model's performance. In the case of the IMDB binary sentiment classification dataset, the model outputs a large number of "neutral" responses, which are not included in the application label space, resulting in a performance drop (Section 4.4.2).

- **Significant improvement in zero/few-shot scenarios:** In zero-shot and few-shot scenarios, GPT-3.5 outperforms existing LLMs in most NLU tasks, especially in reading comprehension, natural language inference, and semantic matching tasks (Table 21).

---
[1]Details of prompts can be found in the Appendix B

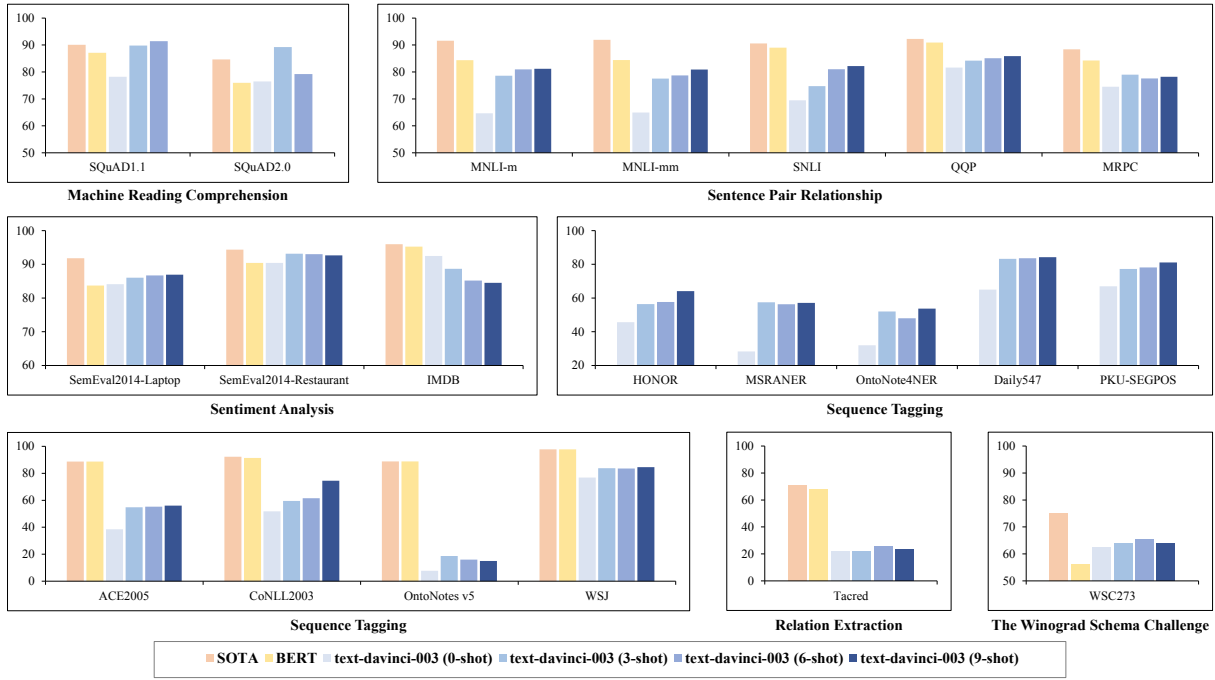

Figure 1: Performance of GPT-3.5 in zero-shot and few-shot (1-shot, 3-shot, 6-shot, 9-shot) scenarios, BERT-base model, and models fine-tuned with task-specific data for 21 different datasets. For each dataset, the better result between all prompts is shown. Details of results can be found in Appendix A.

- **Ability for in-context learning:** Compared to 0-shot, GPT-3.5 performs better on most tasks in the 1-shot setting. Additionally, performance does not vary significantly between the 1-shot, 3-shot, 6-shot, and 9-shot settings for most tasks. However, providing additional examples in the prompts can be advantageous for sequence tagging tasks (Figure 1 and Section 4.5).

- **Differences between GPT models:** In most NLU tasks, GPT-3.5 models (*text-davinci-003*, *text-davinci-002*) demonstrate a significant improvement over GPT-3 model (*text-davinci-001*), particularly in tasks that require a higher level of language understanding, such as reading comprehension, natural language reasoning, and sequence tagging.

Based on the findings above, we arrive at the following three conclusions:

- **Sources of model comprehension ability:** The combination of pretraining, code training, and instruction tuning stages plays a decisive role in enhancing the model's ability to natural language understanding. We speculate that pretraining provides the model with a foundation of semantic understanding, code training enhances its ability to comprehend semantic dependencies, and instruction tuning improves its generalizability to novel tasks.

- **Factors affecting robustness:** The adoption of supervised training during the instruction tuning stage for both instructions and tasks could be a crucial factor leading to robustness issues akin to the pretraining-finetuning paradigm. Therefore, beyond examining the robustness of tasks, it is imperative to conduct further research into the robustness of instructions.

- **Consistency of task labels and label types:** Ensuring the consistency of task labels and their types between the instruction tuning stage and the actual application stage is crucial for the model's performance and generalizability. As such, it's important to design effective prompts or instructions that guarantee this consistency and adapt to different label types and application scenarios.

Table 1: Information of all datasets used in experiments.

| Task | Subtask | Dataset | # Samples | Measure | Language |
|---|---|---|---|---|---|
| Machine Reading Comprehension | Machine Reading Comprehension | SQuAD1.1 | 9868 | F1 & EM | English |
| | | SQuAD2.0 | 11491 | F1 & EM | English |
| Relation Extraction | Relation Extraction | Tacred | 15509 | F1 | English |
| Sentence Pair Relationship | Natural Language Inference | MNLI-m | 9815 | Accuracy | English |
| | | MNLI-mm | 9832 | Accuracy | English |
| | | SNLI | 10000 | Accuracy | English |
| | Semantic Matching | MRPC | 1724 | Accuracy | English |
| | | QQP | 5000 | Accuracy | English |
| Sentiment Analysis | Aspect-based Sentiment Analysis | SemEval2014-Laptop | 331 | Accuracy | English |
| | | SemEval2014-Restaurant | 492 | Accuracy | English |
| | Sentiment Classification | IMDB | 11257 | Accuracy | English |
| Sequence Tagging | Named Entity Recognition | ACE 2005 | 1312 | F1 | English |
| | | CoNLL 2003 | 3453 | F1 | English |
| | | OntoNotes v5 | 4019 | F1 | English |
| | | HONOR | 1120 | F1 | Chinese |
| | | MSRANER | 4365 | F1 | Chinese |
| | | OntoNote4NER | 4346 | F1 | Chinese |
| | Part-of-speech Tagging | Daily547 | 546 | Accuracy | English |
| | | WSJ | 5461 | Accuracy | English |
| | | PKU-SEGPOS | 5204 | F1 | Chinese |
| The Winograd Schema Challenge | The Winograd Schema Challenge | WSC273 | 570 | Accuracy | English |

## 2 Background and Related Work

### 2.1 Large Language Models

In recent years, LLMs have become increasingly popular in natural language processing research. The development of these models has been driven by breakthroughs such as the 175 billion parameters GPT-3 model (Brown et al., 2020), and advancements in the field, including FLAN (Wei et al., 2022), T0 (Sanh et al., 2021), LaMDA (Thoppilan et al., 2022), and PaLM (Chowdhery et al., 2022), have contributed significantly to the natural language processing field. Recently, the release of GPT-3.5 and ChatGPT[2] has garnered significant attention for their detailed responses and articulate answers across many domains of knowledge (Guo et al., 2023; Qin et al., 2023), which are fine-tuned from the GPT-3.5 series with Reinforcement Learning from Human Feedback (RLHF) (Christiano et al., 2017). Despite the enormous potential of these LLMs, previous research (Hendy et al., 2023; Bang et al., 2023) has mainly focused on exploring their performance across various tasks, with little attention paid to their robustness, which can measure their abilities to handle various complexities of the open world and is a crucial dimension of trustworthy AI (Liu et al., 2022). Therefore, in this work, we conduct a comprehensive experimental analysis of GPT-3.5's robustness.

### 2.2 TextFlint

In this work, we leverage TextFlint (Gui et al., 2021), a robustness evaluation toolkit specifically designed for NLP models, to perform a comprehensive robustness evaluation. TextFlint offers a range of text transformations, including universal text transformation, task-specific transformation, adversarial attack, subpopulation, and their combinations, to introduce various types of perturbations to text data. We utilize 66 text transformations from TextFlint for our robustness evaluation.

## 3 Experiment Setup

### 3.1 Task and Datasets

### 3.2 Task and Datasets

We conduct an evaluation of GPT-3.5 using 21 datasets covering 9 popular NLU task categories: machine reading comprehension **(MRC)** (SQuAD1.1(Rajpurkar et al., 2016), SQuAD2.0(Rajpurkar et al., 2018)), relation extraction **(RE)** (Tacred(Zhang et al., 2017)), natural language inference **(NLI)** (MNLI-m(Williams et al., 2017), MNLI-mm(Williams et al., 2017), SNLI(Williams et al., 2017)), semantic matching **(SM)** (MRPC(Dolan and Brockett, 2005), QQP(Wang et al., 2017)), aspect-based sentiment analysis **(ABSA)** (SemEval2014-Laptop(Pontiki et al., 2014), SemEval2014-Restaurant(Pontiki et al., 2014)), sentiment analysis **(SA)** (IMDB(Maas et al., 2011)), named enitity recognition **(NER)** (ACE2005[3], CoNLL2003(Sang and Meulder, 2003), OntoNotesv5[4], HONOR[5], MSRANER (Levow, 2006), OntoNote4NER (Weischedel et al., 2011)), part-of-speech tagginng **(POS)** (Daily547(Gimpel et al., 2010), WSJ(Marcus et al., 1993), PKU-SEGPOS[6] ), and the winograd schema challenge **(WSC)** (WSC273(Levesque et al., 2012)). To assess the model's robustness and performance, we apply a total of 66 text transformations from TextFlint across these tasks. Further details about our experimental dataset can be found in Table 1.

### 3.3 Model Selection

In this study, our main objective is to assess the performance and robustness of the GPT-3.5 models in the aforementioned NLU tasks. To compare the results with earlier GPT models, we specifically chose *text-davinci-002* from the GPT-3.5 series models and *text-davinci-001* from the GPT-3 series models. Additionally, we compare our results with those of existing supervised training models and zero/few-shot models (Section 4.7.2).

### 3.4 Zero/few-shot Scene Setting

For each dataset, we evaluate the models' performance in both zero-shot and few-shot scenarios. We set up four different scenarios: 1/3/6/9-shot. For 1-shot and 3-shot, we test the entire dataset with various task-specific transformations. For 6-shot and 9-shot, we only evaluate 1000 samples in the original dataset to assess the effect of the number of examples in prompts on performance. For further details, please refer to Section 4, Figure 1 and Appendix A.

### 3.5 Prompt Selection

Since LLMs rely on different prompts to perform downstream tasks, we extensively collect a large number of task-specific prompts [7]. We select and design new prompts manually for certain tasks, ultimately choosing the three best-performing prompts per dataset to obtain the most objective results. We expand these three prompts [8] into five different shot levels (0/1/3/6/9-shot) by changing the number of examples in the prompt. Further details about these prompts can be found in Appendix B. Please note that in Section 4, the recorded results for the few-shot scenarios with different numbers of examples represent means and standard deviations of the results obtained from the corresponding three prompts.

## 4 Experiments

In this section, we present the experimental results and conduct an analysis for each task.

---

[2]Despite the lack of the official API for ChatGPT, we plan to continue exploring this model in future research.
[3]https://catalog.ldc.upenn.edu/LDC2006T06
[4]https://catalog.ldc.upenn.edu/LDC2013T19
[5]we construct the HONOR data set ourselves, which focuses on named entity recognition related to scheduling.
[6]http://cuge.baai.ac.cn/\#/dataset?id=19\&name=PKU-SEGPOS
[7]Some of the prompts are sourced from https://github.com/bigscience-workshop/promptsource.
[8]In the RE and ST tasks, since the original labels are abbreviations, we map them to specific phrases in the prompt in order for the model to understand their meaning.

## 4.1 Machine Reading Comprehension

Table 2: Robustness test results of GPT series models (zero-shot and few-shot) and fine-tuned models (i.e., BERT (Devlin et al., 2019), ALBERT (Lan et al., 2019), DistilBERT (Sanh et al., 2019), XLNet (Yang et al., 2019), LUKE (Yamada et al., 2020), SpanBERT (Joshi et al., 2020) and DeBERTa (He et al., 2020)) on **SQuAD1.1** dataset, using micro-F1 score as the evaluation metric.

| Model | AddSentDiverse # 9292 samples | | ModifyPos # 9011 samples | | PerturbAnswer # 9833 samples | | PerturbQuestion-BackTranslation # 9868 samples | | PerturbQuestion-MLM # 9867 samples | |
|---|---|---|---|---|---|---|---|---|---|---|
| | ori | trans | ori | trans | ori | trans | ori | trans | ori | trans |
| *text-davinci-003* | | | | | | | | | | |
| 0-shot | 66.84±9.79 | 55.59±9.95 | 67.55±9.84 | 67.46±8.94 | 67.16±9.71 | 65.97±8.27 | 67.19±9.66 | 59.90±8.00 | 67.18±9.64 | 56.43±9.15 |
| 1-shot | 88.20±1.30 | **70.75±1.98** | 88.47±1.24 | 88.37±1.00 | 88.13±1.27 | 85.27±1.11 | 88.10±1.29 | 80.31±1.19 | 88.13±1.26 | 78.34±1.70 |
| 3-shot | **89.64±0.40** | 68.04±0.85 | 89.94±0.35 | 89.88±0.38 | 89.60±0.37 | 86.92±0.44 | 89.57±0.38 | 82.10±0.55 | 89.57±0.38 | 79.94±0.90 |
| *text-davinci-002* | | | | | | | | | | |
| 0-shot | 78.56±11.01 | 55.67±8.01 | 77.77±11.76 | 77.10±11.19 | 78.52±10.86 | 72.21±10.85 | 78.44±10.84 | 69.04±9.52 | 78.47±10.94 | 64.47±10.43 |
| 1-shot | 58.82±19.61 | 47.43±14.23 | 58.19±19.65 | 57.48±19.94 | 58.42±19.04 | 52.35±15.05 | 58.62±19.10 | 52.43±16.37 | 58.3±18.93 | 49.55±18.01 |
| 3-shot | 82.72±7.87 | 61.97±2.90 | 83.37±8.13 | 82.65±8.01 | 82.80±7.74 | 80.98±7.24 | 82.93±7.66 | 74.96±7.82 | 82.93±7.70 | 71.64±10.81 |
| *text-davinci-001* | | | | | | | | | | |
| 0-shot | 73.38±11.70 | 51.64±14.23 | 72.45±19.65 | 72.53±19.94 | 72.93±19.04 | 65.89±15.05 | 73.17±19.10 | 62.47±16.37 | 73.09±18.93 | 61.27±18.01 |
| 1-shot | 85.01±2.11 | 67.67±0.97 | 84.83±1.74 | 84.28±2.68 | 85.00±2.03 | 79.28±2.11 | 85.11±1.90 | 73.43±1.54 | 85.02±1.90 | 73.02±1.36 |
| 3-shot | 85.41±0.87 | 65.86±1.10 | 85.49±1.31 | 85.33±2.21 | 85.73±0.39 | 80.17±0.61 | 85.30±0.95 | 74.28±0.97 | 85.37±0.85 | 74.81±0.68 |
| *fine-tuned* | | | | | | | | | | |
| BERT | 87.09 | 32.47 | 87.68 | 87.25 | 87.13 | 75.40 | 87.09 | 79.37 | 87.09 | 79.39 |
| ALBERT | 91.27 | 40.45 | 91.76 | 90.82 | 91.26 | 80.52 | 91.23 | 82.73 | 91.23 | 75.59 |
| DistilBERT | 87.10 | 29.60 | 87.56 | 86.69 | 87.04 | 74.92 | 87.00 | 78.50 | 87.01 | 78.93 |
| XLNet | 89.50 | 37.48 | 89.81 | 88.94 | 89.45 | 80.15 | 89.44 | 82.08 | 89.43 | 78.14 |
| LUKE | **95.15** | 57.46 | 95.57 | 94.93 | 95.14 | 87.43 | 95.13 | 87.93 | 95.13 | 76.44 |
| SpanBERT | 94.19 | 50.90 | 94.63 | 94.17 | 94.15 | 86.56 | 94.13 | 86.49 | 94.13 | **84.52** |
| DeBERTa | 93.69 | 46.44 | 94.07 | 93.49 | 93.66 | 85.89 | 93.64 | 86.30 | 93.64 | 75.81 |

Table 3: Robustness test results of GPT series models (zero-shot and few-shot) and fine-tuned models (i.e., BERT, ALBERT, DistilBERT, XLNet, LUKE, SpanBERT and DeBERTa) on **SQuAD1.1** dataset, using EM score as the evaluation metric.

| Model | AddSentDiverse # 1312 samples | | ModifyPos # 1312 samples | | PerturbAnswer # 1405 samples | | PerturbQuestion-BackTranslation # 1312 samples | | PerturbQuestion-MLM # 1312 samples | |
|---|---|---|---|---|---|---|---|---|---|---|
| | ori | trans | ori | trans | ori | trans | ori | trans | ori | trans |
| *text-davinci-003* | | | | | | | | | | |
| 0-shot | 42.76±15.43 | 34.01±15.57 | 43.64±15.58 | 43.71±14.66 | 43.37±15.34 | 42.58±13.32 | 43.37±15.31 | 36.73±12.78 | 43.34±15.31 | 34.05±13.79 |
| 1-shot | 74.51±3.13 | **57.93±3.08** | 75.03±3.00 | 75.26±2.60 | 74.55±3.07 | 70.66±2.65 | 74.47±3.07 | 65.54±2.80 | 74.50±3.04 | 62.86±3.14 |
| 3-shot | **77.85±1.24** | 57.53±0.46 | 78.41±1.15 | 78.61±1.19 | 77.93±1.18 | 74.29±1.19 | 77.86±1.15 | 69.04±1.49 | 77.86±1.19 | 66.23±1.75 |
| *text-davinci-002* | | | | | | | | | | |
| 0-shot | 65.33±15.96 | 43.13±12.76 | 63.57±17.25 | 62.80±16.62 | 65.27±15.61 | 58.07±15.88 | 65.33±15.61 | 55.27±13.57 | 65.30±15.70 | 50.50±13.65 |
| 1-shot | 51.20±19.81 | 40.97±14.12 | 50.53±19.99 | 49.87±20.40 | 50.63±19.27 | 44.57±14.77 | 50.9±19.33 | 44.5±15.96 | 50.47±19.17 | 41.27±18.21 |
| 3-shot | 76.60±8.77 | 56.5±3.50 | 77.23±9.03 | 76.37±8.61 | 76.73±8.78 | 74.77±7.39 | 76.93±8.66 | 67.77±8.14 | 76.87±8.70 | 64.87±11.20 |
| *text-davinci-001* | | | | | | | | | | |
| 0-shot | 58.17±16.43 | 38.40±16.18 | 56.67±16.95 | 57.07±15.46 | 57.63±16.28 | 50.03±15.24 | 57.87±16.34 | 47.87±13.25 | 57.73±16.09 | 47.57±13.37 |
| 1-shot | 73.43±3.39 | 57.43±0.76 | 73.27±2.85 | 73.17±4.00 | 73.43±3.20 | 66.97±2.52 | 73.50±3.02 | 62.07±2.65 | 73.50±3.06 | 62.00±2.26 |
| 3-shot | 76.05±1.41 | 57.78±1.23 | 76.16±2.05 | 76.23±3.42 | 76.5±1.37 | 70.35±1.33 | 75.97±1.74 | 65.00±1.35 | 76.07±1.52 | 65.86±1.05 |
| *fine-tuned* | | | | | | | | | | |
| BERT | 79.25 | 27.93 | 79.95 | 79.81 | 79.30 | 62.48 | 79.25 | 70.16 | 79.24 | 70.38 |
| ALBERT | 84.70 | 35.87 | 85.31 | 84.24 | 84.63 | 68.80 | 84.56 | 74.44 | 84.57 | 61.04 |
| DistilBERT | 79.43 | 25.53 | 79.96 | 79.10 | 79.35 | 62.21 | 79.30 | 69.26 | 79.31 | 70.08 |
| XLNet | 81.37 | 32.12 | 81.79 | 81.13 | 81.30 | 67.15 | 81.23 | 72.72 | 81.23 | 65.55 |
| LUKE | **90.21** | 52.73 | 90.73 | 90.01 | 90.08 | 77.00 | 90.06 | 81.12 | 90.07 | 60.19 |
| SpanBERT | 88.64 | 46.72 | 89.01 | 88.50 | 88.47 | 75.38 | 88.42 | 78.88 | 88.42 | **76.50** |
| DeBERTa | 88.29 | 42.24 | 88.76 | 88.08 | 88.18 | 75.10 | 88.10 | 79.18 | 88.10 | 59.94 |

Table 4: Robustness test results of GPT series models (zero-shot and few-shot) and fine-tuned models (i.e., BERT, XLNet and DeBERTa) on **SQuAD2.0** dataset, using micro-F1 score as the evaluation metric.

| Model | AddSentDiverse # 5129 samples | | ModifyPos # 5053 samples | | PerturbAnswer # 5522 samples | | PerturbQuestion-BackTranslation # 11492 samples | | PerturbQuestion-MLM # 11491 samples | |
|---|---|---|---|---|---|---|---|---|---|---|
| | ori | trans | ori | trans | ori | trans | ori | trans | ori | trans |
| *text-davinci-003* | | | | | | | | | | |
| 0-shot | 65.42±9.22 | 54.21±9.96 | 66.31±9.45 | 66.16±8.65 | 65.87±9.20 | 65.19±7.96 | 65.96±9.17 | 58.09±7.34 | 66.00±9.11 | 54.87±8.62 |
| 1-shot | 87.50±1.26 | **69.80±2.07** | 88.01±1.26 | 87.80±1.14 | 87.34±1.16 | 84.79±1.07 | 87.24±1.30 | 79.16±1.14 | 87.36±1.24 | 77.04±1.67 |
| 3-shot | **88.88±0.38** | 67.08±0.90 | **89.29±0.37** | **89.41±0.47** | **88.83±0.43** | **86.36±0.42** | **88.88±0.36** | **80.89±0.51** | **88.87±0.46** | **78.87±1.03** |
| *text-davinci-002* | | | | | | | | | | |
| 0-shot | 69.31±12.21 | 48.92±14.03 | 69.05±12.28 | 68.67±12.56 | 68.69±12.14 | 63.83±11.56 | 67.36±11.41 | 56.78±9.97 | 67.50±11.23 | 55.52±13.31 |
| 1-shot | 61.16±17.04 | 48.01±12.27 | 61.14±16.83 | 59.61±16.83 | 61.21±16.64 | 53.20±15.28 | 63.73±14.55 | 55.95±13.82 | 63.67±14.47 | 53.01±14.75 |
| 3-shot | 84.44±5.89 | 57.83±2.05 | 83.64±6.19 | 81.63±7.04 | 83.27±5.90 | 77.73±7.09 | 84.83±4.34 | 73.77±5.07 | 84.67±4.69 | 71.42±7.39 |
| *text-davinci-001* | | | | | | | | | | |
| 0-shot | 65.36±13.26 | 49.07±15.05 | 65.22±12.79 | 65.55±12.61 | 64.60±13.07 | 59.56±11.98 | 62.61±12.37 | 52.86±8.30 | 62.67±12.33 | 50.51±10.25 |
| 1-shot | 80.02±1.33 | 63.38±4.22 | 79.96±1.75 | 77.94±1.67 | 79.15±1.28 | 74.07±1.47 | 76.99±1.03 | 63.77±0.97 | 77.00±1.18 | 62.93±1.50 |
| 3-shot | 81.97±1.67 | 60.88±1.33 | 81.53±1.65 | 80.49±2.60 | 80.63±1.58 | 75.58±0.85 | 77.99±1.47 | 66.04±2.09 | 77.94±1.52 | 64.38±0.83 |
| *fine-tuned* | | | | | | | | | | |
| BERT | 80.38 | **34.33** | 81.40 | 79.70 | 80.39 | 54.74 | 75.95 | 67.95 | 75.95 | 64.13 |
| SpanBERT | **89.25** | 30.85 | **89.93** | **88.26** | **89.10** | **67.22** | 87.67 | **77.68** | 87.67 | **68.71** |
| DeBERTa | 87.90 | 28.38 | 88.52 | 86.28 | 87.76 | 63.19 | **87.72** | 69.16 | **87.72** | 56.60 |

Table 5: Robustness test results of GPT series models (zero-shot and few-shot) and fine-tuned models (i.e., BERT, XLNet and SpanBERT) on **SQuAD2.0** dataset, using EM score as the evaluation metric.

| Model | AddSentDiverse # 1312 samples | | ModifyPos # 1312 samples | | PerturbAnswer # 1405 samples | | PerturbQuestion-BackTranslation # 1312 samples | | PerturbQuestion-MLM # 1312 samples | |
|---|---|---|---|---|---|---|---|---|---|---|
| | ori | trans | ori | trans | ori | trans | ori | trans | ori | trans |
| *text-davinci-003* | | | | | | | | | | |
| 0-shot | 40.55±14.61 | 31.78±15.33 | 41.63±15.09 | 41.54±14.10 | 41.34±14.56 | 41.07±12.80 | 41.50±14.42 | 34.33±11.71 | 41.43±14.52 | 31.76±13.05 |
| 1-shot | 73.00±3.19 | **56.06±3.08** | 73.95±3.16 | 74.08±2.88 | 72.73±2.98 | 69.40±2.60 | 72.83±3.42 | 63.54±2.93 | 73.04±3.25 | 60.72±3.07 |
| 3-shot | **76.2±1.29** | 55.7±0.47 | **76.96±1.17** | **77.57±1.43** | **76.29±1.25** | **72.95±1.24** | **76.42±1.17** | **67.01±1.46** | **76.43±1.33** | **64.19±1.94** |
| *text-davinci-002* | | | | | | | | | | |
| 0-shot | 50.73±17.72 | 33.03±17.64 | 50.13±17.98 | 49.70±18.20 | 49.83±17.31 | 44.47±16.12 | 49.04±15.81 | 40.26±14.91 | 49.11±15.68 | 38.00±16.24 |
| 1-shot | 49.87±16.57 | 38.13±12.02 | 49.67±16.31 | 48.27±16.34 | 49.70±16.10 | 41.50±13.64 | 51.51±14.46 | 43.62±13.46 | 51.51±14.33 | 40.40±14.94 |
| 3-shot | 74.87±6.59 | 49.6±2.60 | 73.57±6.83 | 71.57±7.80 | 73.03±6.54 | 66.80±7.30 | 74.07±5.41 | 62.83±6.49 | 73.87±5.64 | 59.19±7.75 |
| *text-davinci-001* | | | | | | | | | | |
| 0-shot | 45.60±18.09 | 31.87±18.18 | 45.37±17.68 | 46.07±17.56 | 44.67±17.72 | 39.73±15.84 | 43.21±16.90 | 35.46±12.67 | 43.21±16.90 | 32.92±12.81 |
| 1-shot | 65.43±2.02 | 50.03±2.97 | 64.93±2.26 | 63.67±2.31 | 63.93±2.05 | 58.83±1.81 | 61.66±1.37 | 49.66±1.73 | 61.59±1.66 | 48.15±1.03 |
| 3-shot | 69.75±2.42 | 49.47±0.67 | 68.91±2.49 | 68.27±3.15 | 67.87±2.22 | 62.97±1.45 | 65.16±1.38 | 54.44±1.99 | 65.09±1.73 | 51.45±1.10 |
| *fine-tuned* | | | | | | | | | | |
| BERT | 73.66 | **30.55** | 74.73 | 72.93 | 73.63 | 45.89 | 72.67 | 64.68 | 72.67 | 61.21 |
| SpanBERT | **83.04** | 28.15 | **83.81** | **82.27** | **82.87** | **58.31** | **84.62** | **74.55** | **84.62** | **66.01** |
| DeBERTa | 82.30 | 26.17 | 82.90 | 80.69 | 82.00 | 55.00 | 81.88 | 63.40 | 81.90 | 41.53 |

For the MRC task, we compare the micro-F1 and EM scores of 0/1/3-shot GPT series models and fine-tuned models under different robustness tests on SQuAD1.1 and SQuAD2.0 datasets (Table 2 ∼ 5). From the table, we can observe the following characteristics. **Few-shot models are comparable to fine-tuned models in terms of performance and robustness.** This result indicates that few-shot learning can improve the model's robustness while maintaining high performance. Notably, GPT series models display significantly better robustness than the fine-tuned models for AddSentDiverse and PerturbAnswer transformations on SQuAD1.1. **The number of examples in the prompt has inconsistent effects on robustness and performance.** The robustness of 0/1/3-shot models remaine relatively unchanged, but the micro-F1 and EM scores of 1/3-shot models improve significantly compared to 0-shot models. This suggests that few-shot instances can help models learn the current reading comprehension patterns and enhance their reading comprehension abilities for this task. Few-shot models also exhibit less variance between different prompts and are less affected by changes in data and prompts, highlighting their stability in the MRC task.

## 4.2 Relation Extraction

Table 6: Robustness test results of GPT series models (zero-shot and few-shot) and fine-tuned models (i.e., BERT-base-uncased (Devlin et al., 2019), SpanBERT and NLL (Zhou and Chen, 2021)) on **Tacred** dataset.

(a)

| Model | InsertClause #14897 samples | | SwapEnt-LowFreq #15509 samples | | SwapEnt-MultiType #15509 samples | | SwapEnt-SamEtype #15509 samples | |
|---|---|---|---|---|---|---|---|---|
| | ori | trans | ori | trans | ori | trans | ori | trans |
| *text-davinci-003* | | | | | | | | |
| 0-shot | 20.37±1.32 | 18.76±0.88 | 20.40±1.33 | 21.77±1.58 | 20.36±1.24 | 20.23±1.02 | 20.36±1.28 | 22.01±1.63 |
| 1-shot | **23.20±1.63** | **22.22±1.21** | **23.24±1.71** | **23.65±1.72** | **23.25±1.68** | **21.91±1.44** | **23.22±1.72** | **24.06±1.70** |
| 3-shot | 22.01±0.11 | 20.89±0.34 | 21.98±0.18 | 21.53±0.29 | 21.94±0.17 | 19.24±0.73 | 21.96±0.20 | 22.14±0.20 |
| *text-davinci-002* | | | | | | | | |
| 0-shot | 12.34±0.30 | 11.50±0.43 | 12.23±0.09 | 9.19±0.69 | 12.18±0.17 | 9.29±0.81 | 12.23±0.09 | 10.70±0.17 |
| 1-shot | 17.25±0.67 | 14.68±0.68 | 17.03±0.61 | 15.14±0.13 | 17.20±0.61 | 12.50±0.10 | 17.20±0.45 | 15.02±0.87 |
| 3-shot | 17.13±1.14 | 14.48±1.36 | 17.14±1.14 | 14.81±0.35 | 16.85±1.48 | 12.21±1.13 | 17.19±1.33 | 16.11±1.53 |
| *text-davinci-001* | | | | | | | | |
| 0-shot | 10.29±0.73 | 9.27±0.75 | 10.05±0.88 | 11.08±0.91 | 10.15±0.73 | 9.88±1.09 | 10.18±0.81 | 10.43±0.35 |
| 1-shot | 10.57±0.61 | 10.16±0.93 | 10.55±0.60 | 11.62±1.55 | 10.66±0.66 | 10.25±2.08 | 10.50±0.56 | 11.68±1.45 |
| 3-shot | 10.10±0.42 | 9.48±0.19 | 10.15±0.33 | 10.48±0.72 | 9.87±0.19 | 7.44±0.19 | 9.99±0.34 | 10.05±0.66 |
| *fine-tuned* | | | | | | | | |
| BERT-base-uncased | 68.01 | 50.95 | 68.01 | 59.16 | 68.01 | 48.66 | 68.01 | 61.91 |
| SpanBERT | 70.81 | 55.79 | 70.81 | 70.81 | **70.81** | **70.81** | 70.81 | 70.81 |
| NLL | **70.94** | **57.13** | **70.87** | **70.87** | - | - | **70.87** | **70.87** |

(b)

| Model | SwapTriplePos-Age #28 samples | | SwapTriplePos-Birth #48 samples | | SwapTriplePos-Employee #251 samples | |
|---|---|---|---|---|---|---|
| | ori | trans | ori | trans | ori | trans |
| *text-davinci-003* | | | | | | |
| 0-shot | 96.43±0.00 | **100.00±0.00** | 65.28±1.21 | 71.53±1.21 | 6.53±2.36 | 16.37±3.96 |
| 1-shot | **100.00±0.00** | **100.00±0.00** | 62.74±5.92 | 68.31±2.83 | 10.88±1.07 | 16.51±2.26 |
| 3-shot | **100.00±0.00** | **100.00±0.00** | 75.00±2.08 | 81.25±0.00 | 19.27±3.72 | 25.02±4.99 |
| *text-davinci-002* | | | | | | |
| 0-shot | 89.29±6.19 | 88.10±2.07 | 56.25±8.33 | 52.08±4.17 | 16.47±7.02 | 17.40±8.52 |
| 1-shot | 96.61±0.31 | **100.00±0.00** | 62.50±2.08 | 65.97±3.18 | 47.54±6.79 | 52.32±6.68 |
| 3-shot | **100.00±0.00** | **100.00±0.00** | 65.97±13.39 | 63.89±7.89 | **60.03±9.42** | **62.02±6.86** |
| *text-davinci-001* | | | | | | |
| 0-shot | **100.00±0.00** | **100.00±0.00** | 33.90±11.69 | 36.64±7.31 | 0.41±0.41 | 1.22±1.23 |
| 1-shot | **100.00±0.00** | **100.00±0.00** | 32.05±9.63 | 35.26±10.15 | 1.44±1.46 | 1.78±1.47 |
| 3-shot | **100.00±0.00** | **100.00±0.00** | 47.91±9.55 | 48.61±7.89 | 0.80±1.38 | 0.66±1.15 |
| *fine-tuned* | | | | | | |
| BERT-base-uncased | **100.00** | 92.31 | 82.11 | 67.42 | 91.75 | **78.28** |
| SpanBERT | **100.00** | 98.18 | 95.83 | **94.74** | **93.14** | 77.78 |
| NLL | **100.00** | **100.00** | **96.84** | 93.62 | - | - |

For the RE task, we use Tacred as the test dataset and show the micro-f1 scores under different robustness tests in Table 6. From the table, we can have the following observations. **The GPT series models do not seem to be sufficiently capable of solving complex semantic understanding tasks like RE.** Specifically, the fine-tuned models can achieve 70% micro-F1 scores when the data size is large, but the best performance of the GPT series models is only about 20%, which makes it almost impossible to select them for RE tasks. **However, the GPT series models have outstanding performance and robustness for some specific relational judgments.** In particular, for the relationship of person and age (i.e., SwapTriplePos-Age), the GPT series models perform very accurately, and even better for the recognition of transformed sentences. Upon analysis, we find that the syntax of the sentences in the dataset for such specific relations is very fixed, and the transformed sentences often correspond to conventional expressions, providing the models with hints that enable them to make predictions.

## 4.3 Sentence Pair Relationship

We test the sentence pair relationship task on two subtasks: NLI (Section 4.3.1) and SM (Section 4.3.2).

### 4.3.1 Natural Language Inference

Table 7: Robustness test results of GPT series models (zero-shot and few-shot) and fine-tuned (i.e., BERT-base-uncased, RoBERTa-large (Liu et al., 2019), ALBERT-xxlarge-v2 (Lan et al., 2019), DeBERTa-v2-xxlarge (He et al., 2020) ) on **MNLI-m** dataset.

| Model | AddSent # 9815 samples | | NumWord # 745 samples | | SwapAnt # 199 samples | |
|---|---|---|---|---|---|---|
| | ori | trans | ori | trans | ori | trans |
| *text-davinci-003* | | | | | | |
| 0-shot | 64.26±0.53 | 34.04±1.85 | 68.12±1.42 | **51.62±13.17** | 70.11±1.57 | 78.96±6.20 |
| 1-shot | **73.14±4.80** | 41.07±5.05 | **74.43±4.90** | 43.80±3.61 | 93.35±1.66 | 72.87±5.45 |
| 3-shot | 72.07±5.69 | 41.02±3.14 | 71.59±6.01 | 46.49±4.12 | **98.66±0.58** | 82.41±5.10 |
| *text-davinci-002* | | | | | | |
| 0-shot | 52.62±6.64 | 36.27±1.61 | 52.72±5.83 | 29.21±14.81 | 52.27±15.07 | 50.95±19.87 |
| 1-shot | 68.60±6.20 | 37.20±1.10 | 69.44±9.26 | 45.37±14.51 | 90.29±1.05 | 86.26±10.18 |
| 3-shot | 70.30±5.47 | 36.87±2.12 | 71.18±6.00 | 44.61±11.68 | 97.49±1.51 | **88.27±12.51** |
| *text-davinci-001* | | | | | | |
| 0-shot | 42.20±4.15 | 36.66±2.06 | 38.91±3.03 | 25.76±8.35 | 42.41±23.67 | 28.89±23.00 |
| 1-shot | 39.90±8.01 | 39.40±7.63 | 40.44±7.19 | 1.39±2.40 | 98.16±3.19 | 4.36±6.28 |
| 3-shot | 48.80±3.64 | **43.87±3.91** | 50.16±6.62 | 4.34±5.34 | 91.79±7.33 | 28.64±16.34 |
| *fine-tuned* | | | | | | |
| BERT-base-uncased | 84.39 | 55.32 | 83.09 | 49.40 | 91.46 | 52.76 |
| RoBERTa-large | 90.61 | 70.95 | 88.59 | **53.69** | 93.47 | 70.35 |
| ALBERT-xxlarge-v2 | 89.45 | **79.74** | 86.85 | 49.53 | 90.95 | 68.84 |
| DeBERTa-v2-xxlarge | **91.60** | 71.68 | **91.14** | 48.46 | 92.46 | **78.89** |

Table 8: Robustness test results of GPT series models (zero-shot and few-shot) and fine-tuned models (i.e., BERT-base-uncased, RoBERTa-large, ALBERT-xxlarge-v2, DeBERTa-v2-xxlarge ) on **MNLI-mm** dataset.

| Model | AddSent # 9832 samples | | NumWord # 775 samples | | SwapAnt # 255 samples | |
|---|---|---|---|---|---|---|
| | ori | trans | ori | trans | ori | trans |
| *text-davinci-003* | | | | | | |
| 0-shot | 64.56±0.32 | 34.26±2.15 | 67.50±0.25 | 39.81±3.04 | 67.76±0.54 | 77.65±2.71 |
| 1-shot | 70.28±5.21 | 37.01±3.34 | 71.39±4.39 | 28.45±1.80 | 93.20±2.94 | 66.80±6.64 |
| 3-shot | **73.66±3.44** | 37.92±1.28 | **74.99±2.36** | 41.59±4.71 | **96.34±1.20** | 89.61±2.90 |
| *text-davinci-002* | | | | | | |
| 0-shot | 49.93±6.44 | 36.05±2.27 | 52.93±7.57 | 28.45±14.29 | 51.40±15.02 | 58.39±23.61 |
| 1-shot | 51.12±18.88 | 35.20±6.03 | 52.96±20.00 | 16.95±7.03 | 55.30±36.10 | 51.29±18.91 |
| 3-shot | 61.81±11.09 | 34.79±0.70 | 61.40±14.05 | **49.71±20.48** | 70.67±19.95 | 83.12±26.82 |
| *text-davinci-001* | | | | | | |
| 0-shot | 44.07±3.54 | 35.72±3.46 | 45.34±6.09 | 28.11±9.66 | 33.42±18.59 | 31.07±18.61 |
| 1-shot | 47.54±2.85 | 44.87±2.63 | 45.22±2.84 | 11.10±5.54 | 91.76±8.38 | 36.69±6.32 |
| 3-shot | 54.43±3.50 | **49.03±10.00** | 52.16±5.08 | 7.87±10.58 | 89.41±9.90 | 32.29±23.66 |
| *fine-tuned* | | | | | | |
| BERT-base-uncased | 84.43 | 55.25 | 82.97 | 46.06 | 85.10 | 55.69 |
| RoBERTa-large | 90.12 | 67.75 | 88.65 | **52.90** | 92.16 | 74.90 |
| ALBERT-xxlarge-v2 | 89.89 | **79.10** | 89.03 | 44.77 | 91.76 | 69.80 |
| DeBERTa-v2-xxlarge | **91.90** | 72.50 | **91.74** | 47.23 | **94.51** | 76.08 |

Table 9: Robustness test results of GPT series models (zero-shot and few-shot) and fine-tuned models (i.e., BERT-base-uncased, BERT-large-uncased (Devlin et al., 2019), XLNet-large-cased (Yang et al., 2019)) on **SNLI** dataset.

| Model | AddSent # 10000 samples | | NumWord # 108 samples | | SwapAnt # 523 samples | |
|---|---|---|---|---|---|---|
| | ori | trans | ori | trans | ori | trans |
| *text-davinci-003* | | | | | | |
| 0-shot | 67.16±3.26 | 34.11±0.03 | 63.89±2.45 | 73.77±22.76 | 81.39±8.06 | 61.44±17.44 |
| 1-shot | 71.81±4.12 | 34.63±0.26 | 73.30±1.63 | 54.63±6.68 | **97.77±0.67** | 49.46±8.36 |
| 3-shot | 72.18±2.37 | 40.00±2.68 | **75.00±3.34** | 67.29±7.54 | 94.65±2.20 | 45.41±7.14 |
| *text-davinci-002* | | | | | | |
| 0-shot | 47.40±8.24 | 35.45±2.04 | 45.98±11.81 | 30.84±25.56 | 58.76±41.98 | 27.89±20.28 |
| 1-shot | 72.20±3.04 | **40.70±3.72** | 71.60±2.98 | 46.91±10.65 | 94.45±4.45 | 60.48±13.19 |
| 3-shot | **72.43±3.11** | 37.83±2.73 | 70.68±2.97 | 63.27±15.28 | 94.39±3.75 | 66.35±11.72 |
| *text-davinci-001* | | | | | | |
| 0-shot | 38.61±7.26 | 33.00±3.80 | 48.3±26.79 | 46.42±30.83 | 24.75±15.46 | 63.38±37.26 |
| 1-shot | 36.27±0.31 | 34.50±0.61 | 41.97±2.33 | 62.66±11.50 | 2.10±3.48 | 38.24±22.93 |
| 3-shot | 44.60±4.45 | 34.10±0.78 | 47.33±6.43 | **87.94±11.21** | 41.27±26.07 | **89.68±6.84** |
| *fine-tuned* | | | | | | |
| BERT-base-uncased | 88.99 | **79.66** | 91.67 | 58.33 | **91.59** | 94.84 |
| BERT-large-uncased | 89.37 | 58.34 | 91.67 | 67.59 | 90.63 | 94.84 |
| XLNet-large-cased | **90.60** | 72.97 | 91.01 | **91.67** | 62.96 | **95.79** |

We test on MNLI-m, MNLI-mm, SNLI datasets for the NLI task and obtain Table 7, 8, 9. The evaluation reveals that **the robustness varies in different kinds of transformation, and the selection of number of examples in the prompt also has some effect on the results.** For example, for the MNLI-m and MNLI-mm datasets (Table 7 and Table 8), the AddSent transformation shows slightly lower robustness in the 0/1/3-shot models compared to the fine-tuned models, while the NumWord transformation has comparable robustness. Interestingly, the SwapAnt transformation demonstrates significantly better robustness for the 0/1/3-shot models than the fine-tuned models.

### 4.3.2 Semantic Matching

Table 10: Robustness test results of GPT series models (zero-shot and few-shot) and fine-tuned models (i.e., BERT-base-uncased, Funnel-Transformer-medium (Dai et al., 2020) and DeBERTa-base (He et al., 2020)) on **MRPC** dataset.

| Model | all # 1724 samples | NumWord # 402 samples | | SwapAnt # 158 samples | |
|---|---|---|---|---|---|
| | ori | ori | trans | ori | trans |
| *text-davinci-003* | | | | | |
| 0-shot | 70.17±4.51 | **74.63±1.97** | 94.44±3.90 | 75.11±8.26 | 54.22±10.41 |
| 1-shot | 69.50±5.41 | 69.82±5.31 | 98.26±1.88 | 62.02±12.80 | 69.62±4.56 |
| 3-shot | 68.50±10.26 | 67.99±7.34 | 97.51±1.63 | 60.55±17.15 | 68.78±7.42 |
| *text-davinci-002* | | | | | |
| 0-shot | **72.73±2.55** | 68.41±6.24 | 66.67±35.79 | **95.57±5.18** | 36.29±18.66 |
| 1-shot | 69.57±8.35 | 72.31±7.04 | **98.59±1.65** | 64.14±14.24 | 78.69±1.93 |
| 3-shot | 72.70±3.57 | 73.14±2.60 | 96.10±6.53 | 66.45±5.80 | **85.86±9.69** |
| *text-davinci-001* | | | | | |
| 0-shot | 21.60±19.25 | 17.58±21.19 | 17.08±17.04 | 22.79±27.59 | 13.71±11.91 |
| 1-shot | 70.40±1.87 | 65.84±3.45 | 78.44±8.76 | 89.66±2.86 | 49.16±7.63 |
| 3-shot | 53.80±4.41 | 56.05±2.91 | 98.42±1.12 | 50.00±9.74 | 75.11±5.74 |
| *fine-tuned* | | | | | |
| BERT-base-uncased | 84.28 | **91.14** | 0.00 | 82.59 | 0.00 |
| Funnel-Transformer-medium | 87.07 | **91.14** | 1.90 | 86.32 | 0.25 |
| DeBERTa-base | **88.40** | **91.14** | 0.00 | **88.31** | 0.50 |

Table 11: Robustness test results of GPT series models (zero-shot and few-shot) and fine-tuned models (i.e., BERT-base-uncased, ALBERT-xxlarge-v2 and Electra-base(Clark et al., 2020) ) on **QQP** dataset.

| Model | all<br># 5000 samples<br>ori | NumWord<br># 2607 samples<br>ori | | SwapAnt<br># 883 samples<br>ori | |
|---|---|---|---|---|---|
| | | ori | trans | ori | trans |
| *text-davinci-003* | | | | | |
| 0-shot | 81.03±0.67 | 79.85±1.37 | 73.22±19.84 | 60.44±8.78 | 65.31±5.15 |
| 1-shot | 80.93±1.91 | 79.81±1.69 | 73.98±20.42 | 63.42±9.92 | 51.34±7.89 |
| 3-shot | **82.97±1.72** | **83.35±0.14** | 71.06±15.50 | 71.46±9.13 | 56.21±8.66 |
| *text-davinci-002* | | | | | |
| 0-shot | 63.00±5.69 | 68.07±5.70 | 25.00±19.56 | **85.32±13.47** | 28.43±20.73 |
| 1-shot | 77.70±3.05 | 79.50±7.11 | **90.17±24.47** | 47.41±12.95 | 67.87±6.12 |
| 3-shot | 80.30±1.15 | 82.60±6.87 | 87.13±19.29 | 57.53±14.28 | 75.27±6.91 |
| *text-davinci-001* | | | | | |
| 0-shot | 36.40±5.82 | 35.37±2.99 | 6.07±9.48 | 75.65±23.45 | 14.01±19.48 |
| 1-shot | 66.50±3.35 | 66.17±3.43 | 86.47±11.10 | 26.61±23.80 | **90.79±11.85** |
| 3-shot | 65.17±5.29 | 72.60±6.48 | 72.9±12.68 | 67.01±6.18 | 76.48±9.44 |
| *fine-tuned* | | | | | |
| BERT-base-uncased | 90.91 | 89.92 | 48.58 | 93.59 | 56.96 |
| ALBERT-xxlarge-v2 | **92.28** | 91.05 | 68.40 | **94.51** | 53.47 |
| Electra-base | 91.72 | **91.39** | **69.99** | 93.86 | **68.93** |

We test on MRPC and QQP datasets for the SM task and obtain Table 10 and Table 11. From the table, we observe some interesting phenomena. **GPT3.5 has significantly improved robustness compared to the fine-tuned models, as evidenced by the results of the NumWord and SwapAnt transformations.** The 0-shot, 1-shot, and 3-shot models show significantly better robustness than the fine-tuned models, especially on the MRPC dataset (Table 10). It could be because different prompts tend to produce different outputs. Under specific transformation conditions, such as SwapAnt transformation, samples that are original all entailment become all contradictory, resulting in such results. **However, the effect of different transformations and instance numbers on this dataset is not significant.** The robustness of 1-shot and 3-shot models is better than 0-shot. Moreover, there is little difference in robustness between 1-shot and 3-shot, indicating that providing example pairs can effectively improve the model's robustness on this dataset, but the number of examples does not significantly impact the results. The performance improvement of 1-shot and 3-shot models are significant in the NumWord transformed data on the MRPC dataset, reaching 100% accuracy. It is worth mentioning that **the GPT series models show significantly better robustness compared to fine-tuned models on the MRPC dataset.** Fine-tuned models have poor robustness on NumWord and SwapAnt variations in the MRPC dataset, with the accuracy of the transformed data often close to 0. However, neither the text-davinci-003, 002 nor 001 models exhibit such problems. This suggests that the GPT series models are more capable of handling text variations and exceptional cases on this task.

### 4.4 Sentiment Analysis

We specifically test the sentiment analysis task on two subtasks: ABSA (Section 4.4.1) and SC (Section 4.4.2).

### 4.4.1 Aspect-based Sentiment Analysis

Table 12: Robustness test results of GPT series models (zero-shot and few-shot) and fine-tuned models (i.e., BERT-base-uncased, LCF-BERT(Zhang et al., 2019), BERT-aspect (Devlin et al., 2019), AEN-BERT(Song et al., 2019) and Microsoft (API)) on **SemEval2014-Laptop** dataset.

| Model | AddDiff # 331 samples | | ReverseNonTarget # 104 samples | | ReverseTarget # 331 samples | |
|---|---|---|---|---|---|---|
| | ori | trans | ori | trans | ori | trans |
| *text-davinci-003* | | | | | | |
| 0-shot | 83.84±0.33 | 77.50±2.43 | 82.43±0.42 | 39.61±4.83 | 83.62±0.12 | 47.04±4.64 |
| 1-shot | 85.77±0.69 | 87.17±4.62 | 85.63±1.05 | 52.22±8.86 | 85.84±0.57 | 57.07±7.43 |
| 3-shot | 85.91±0.12 | **88.73±4.11** | 85.08±0.48 | **55.59±11.11** | 85.98±0.12 | **59.14±7.11** |
| *text-davinci-002* | | | | | | |
| 0-shot | **86.38±0.11** | 81.90±0.35 | 85.57±0.21 | 52.97±2.96 | **86.4±0.26** | 56.68±5.17 |
| 1-shot | 86.05±0.43 | 82.20±1.98 | 85.22±0.24 | 55.18±2.38 | 86.05±0.43 | 56.77±3.08 |
| 3-shot | 85.41±0.43 | 81.55±1.86 | 84.80±0.97 | 54.01±2.28 | 85.48±0.33 | 56.65±2.43 |
| *text-davinci-001* | | | | | | |
| 0-shot | 85.21±1.70 | 80.10±2.11 | **85.89±1.68** | 47.35±4.16 | 85.26±2.17 | 53.56±0.92 |
| 1-shot | 84.56±2.59 | 77.55±4.34 | 84.64±3.34 | 45.02±1.26 | 84.76±2.80 | 51.11±5.13 |
| 3-shot | 83.33±0.69 | 71.90±0.12 | 83.40±0.24 | 48.40±1.98 | 83.44±0.87 | 50.26±0.77 |
| *fine-tuned* | | | | | | |
| BERT-base-uncased | 83.69 | 70.82 | 85.77 | 43.51 | 83.69 | 46.14 |
| LCF-BERT | 81.97 | 77.90 | 82.85 | 56.90 | 81.97 | 48.93 |
| BERT-aspect | 81.97 | 72.96 | 82.85 | 51.05 | 81.97 | 54.08 |
| AEN-BERT | 81.55 | 71.89 | 83.26 | 40.17 | 81.55 | 50.64 |
| Microsoft (API) | **91.79** | **90.63** | **91.79** | **66.06** | **91.79** | **69.70** |

Table 13: Robustness test results of GPT series models (zero-shot and few-shot) and fine-tuned models (i.e., BERT-base-uncased, LCF-BERT, BERT-aspect, AEN-BERT and Microsoft (API)) on **SemEval2014-Restaurant** dataset.

| Model | AddDiff # 492 samples | | ReverseNonTarget # 227 samples | | ReverseTarget # 492 samples | |
|---|---|---|---|---|---|---|
| | ori | trans | ori | trans | ori | trans |
| *text-davinci-003* | | | | | | |
| 0-shot | 89.45±0.87 | 55.25±20.03 | 91.58±0.96 | 47.93±0.45 | 89.38±0.85 | 54.60±7.18 |
| 1-shot | 92.08±0.74 | 78.30±17.09 | 92.84±0.52 | 64.97±11.58 | 92.05±0.78 | 66.62±8.78 |
| 3-shot | **92.56±0.83** | **89.47±7.53** | **93.01±0.52** | **70.95±5.76** | **92.52±0.78** | 70.68±4.60 |
| *text-davinci-002* | | | | | | |
| 0-shot | 91.86±0.79 | 74.05±19.15 | 92.00±0.91 | 69.50±3.12 | 91.90±0.79 | 69.96±5.18 |
| 1-shot | 92.48±0.54 | 86.46±7.66 | 92.39±0.48 | 68.94±2.88 | 92.41±0.61 | **72.39±4.25** |
| 3-shot | 90.36±0.07 | 74.14±16.42 | 90.84±0.50 | 56.82±4.08 | 90.36±0.07 | 52.73±13.11 |
| *text-davinci-001* | | | | | | |
| 0-shot | 89.25±0.94 | 54.56±12.55 | 90.07±1.24 | 63.35±1.58 | 88.89±1.45 | 63.40±1.94 |
| 1-shot | 88.07±3.09 | 33.70±4.97 | 88.49±3.02 | 61.90±1.22 | 88.03±3.05 | 58.03±3.50 |
| 3-shot | 89.49±0.21 | 50.57±3.50 | 90.43±0.52 | 62.07±1.04 | 89.57±0.25 | 60.73±1.02 |
| *fine-tuned* | | | | | | |
| BERT-base-uncased | 90.44 | 55.37 | 90.55 | 53.09 | 90.44 | 38.72 |
| LCF-BERT | 90.32 | **85.83** | 90.55 | 61.17 | 90.32 | 54.78 |
| BERT-aspect | 90.32 | 81.23 | 91.41 | 56.19 | 90.32 | 64.70 |
| AEN-BERT | 90.20 | 59.86 | 90.21 | 58.08 | 90.20 | 34.59 |
| Microsoft (API) | **94.36** | 82.20 | **94.36** | 69.82 | **94.36** | **75.95** |

We test on SemEval2014-Laptop and SemEval2014-Restaurant for ABSA, as shown in Table 12 and 13. From the tables, we can see that: **1) the 0/1/3-shot models perform well and are comparable to the fine-tuned model.** The models show consistent performance and robustness across the two datasets, and there is little variation in performance between the different prompts. **2) in-context learning has been demonstrated to significantly enhance the model's robustness and ability to handle transformed data.** Compared to the fine-tuned models, the 0-shot model exhibits lower robustness; however, the 1-shot and 3-shot models demonstrate comparable and better robustness, respectively.

### 4.4.2 Sentiment Classification

Table 14: Robustness test results of GPT series models (zero-shot and few-shot) and fine-tuned models (i.e., XLNet, ULMFIT(Howard and Ruder, 2018), BERT-large-ITPT (Sun et al., 2019) and DistilBERT) on **IMDB** dataset.

| Model | AddSum-Movie # 11257 samples | | AddSum-Person # 12230 samples | | DoubleDenial # 22933 samples | | SwapSpecialEnt-Movie # 11257 samples | | SwapSpecialEnt-Person # 12230 samples | |
|---|---|---|---|---|---|---|---|---|---|---|
| | ori | trans | ori | trans | ori | trans | ori | trans | ori | trans |
| *text-davinci-003* | | | | | | | | | | |
| 0-shot | 91.51±1.14 | 91.56±0.80 | 91.59±1.04 | 89.62±0.57 | 92.02±0.77 | 91.13±0.69 | 91.53±1.16 | 91.17±0.77 | 91.60±0.97 | 91.62±0.86 |
| 1-shot | 91.67±0.61 | 91.4±0.66 | 91.99±0.53 | **89.86±0.83** | 92.24±0.50 | 90.60±0.75 | 91.68±0.62 | 91.22±0.46 | 92.00±0.51 | **91.87±0.57** |
| 3-shot | 85.73±3.61 | 84.57±3.83 | 86.22±3.45 | 82.88±3.38 | 87.38±3.10 | 85.80±3.79 | 85.69±3.58 | 85.08±3.82 | 86.23±3.44 | 86.06±3.48 |
| *text-davinci-002* | | | | | | | | | | |
| 0-shot | 91.97±1.27 | 91.17±2.66 | 92.00±1.59 | 87.67±1.85 | **93.33±0.96** | **92.57±1.11** | 91.80±1.41 | 90.90±1.91 | 91.97±1.63 | 91.43±1.97 |
| 1-shot | 89.53±2.20 | 87.10±3.94 | 89.33±2.03 | 80.93±6.54 | 91.67±1.16 | 89.87±1.91 | 89.53±1.78 | 88.50±2.38 | 89.27±2.05 | 88.70±2.75 |
| 3-shot | 87.30±0.62 | 84.77±1.01 | 88.53±1.08 | 76.57±2.74 | 91.33±0.90 | 90.23±0.40 | 87.63±0.68 | 86.53±1.25 | 88.37±1.11 | 88.17±0.90 |
| *text-davinci-001* | | | | | | | | | | |
| 0-shot | **92.63±0.54** | 91.59±0.86 | **92.34±0.38** | 68.41±18.64 | 93.20±0.44 | 76.52±15.39 | **92.53±0.47** | 83.91±13.67 | **92.43±0.42** | 81.96±9.59 |
| 1-shot | 91.93±0.06 | 90.57±0.50 | 91.53±0.51 | 86.26±0.78 | 93.13±0.46 | 92.43±0.57 | 91.57±0.15 | **91.57±0.42** | 91.50±0.35 | 91.33±0.15 |
| 3-shot | 91.13±0.49 | 89.03±0.32 | 91.40±0.82 | 86.08±0.91 | 93.23±0.49 | 92.40±0.26 | 91.00±0.56 | 90.93±0.45 | 91.77±0.58 | 91.83±0.50 |
| *fine-tuned* | | | | | | | | | | |
| XLNet | **95.97** | **95.38** | **95.92** | 93.16 | **96.27** | **95.34** | **95.97** | **95.75** | **95.92** | **95.85** |
| ULMFIT | 94.38 | 94.03 | 94.76 | 92.93 | 94.72 | 89.55 | 94.38 | 94.33 | 94.76 | 94.70 |
| BERT-large-ITPT | 95.12 | 94.97 | 95.27 | **94.39** | 95.40 | 93.30 | 95.12 | 95.14 | 95.27 | 95.23 |
| DistilBERT | 92.00 | 90.65 | 92.35 | 86.30 | 92.80 | 88.92 | 92.00 | 91.91 | 92.35 | 92.38 |

We test the robustness of the SC task on the IMDB dataset and obtain Table 14. **The 0/1/3-shot models show good robustness and are comparable to the fine-tuned models**, with low variance between different prompts. However, GPT3.5 exhibits significant defects in this binary classification dataset, with a "neutral" answer given by models instead of the specified positive or negative sentiment in the prompt. As the number of examples in prompts increased, the proportion of neutral answers also increased, resulting in a decline in 3-shot performance. This may be due to the model encountering three-classification data during the fine-tuning stage, which could affect the evaluation of binary classification data. **Furthermore, shorter examples in prompt seem to be more beneficial for IMDB dateset.** When the selected example length is shorter in the prompt, the performance of the few-shot model is better than that of a long one. It is because when the input prompt is too long, it may negatively affect the model's modeling ability.

### 4.5 Sequence Tagging

We evaluate the robustness and performance of GPT series models on two common sequence tagging tasks, NER and POS. In Section 4.5.1 and Section 4.5.2, we will introduce the experimental results of these two tasks, respectively.

### 4.5.1 Named Entity Recognition

Table 15: Robustness test results of GPT series models (zero-shot and few-shot) and fine-tuned models (i.e., BERT-base-cased (Devlin et al., 2019), BERT-base-uncased, ELMo-BiLSTM-CRF (Peters et al., 2018), Flair Embeddings (Akbik et al., 2018), CNN-BiLSTM-CRF (Ma and Hovy, 2016), GRN (Chen et al., 2019)) on **ACE2005** dataset.

| Model | ConcatSent #1312 samples | | CrossCatagory #1312 samples | | EntTypos #1405 samples | | OOV #1312 samples | | SwapLonger #1312 samples | |
|---|---|---|---|---|---|---|---|---|---|---|
| | ori | trans | ori | trans | ori | trans | ori | trans | ori | trans |
| *text-davinci-003* | | | | | | | | | | |
| 0-shot | 36.93±1.86 | 29.43±1.99 | 37.03±1.69 | 36.33±1.17 | 31.31±2.35 | 26.42±2.31 | 36.97±1.91 | 69.86±2.92 | 36.90±1.85 | 75.39±2.20 |
| 1-shot | 43.37±1.52 | 34.68±1.61 | 43.48±1.48 | 34.94±0.45 | 37.63±1.28 | 30.82±1.32 | 43.43±1.41 | 70.35±1.05 | 43.34±1.49 | 77.42±0.36 |
| 3-shot | **54.11±0.60** | **47.30±1.01** | 53.99±0.59 | 39.21±0.26 | 45.51±0.44 | 43.10±0.24 | 54.29±0.51 | 71.46±0.22 | 54.17±0.77 | 77.44±0.37 |
| *text-davinci-002* | | | | | | | | | | |
| 0-shot | 35.50±1.64 | 37.95±11.49 | 32.65±3.89 | 30.58±1.64 | 37.13±0.86 | 35.22±1.36 | 35.55±1.57 | 59.20±1.87 | 32.60±3.97 | 65.36±1.23 |
| 1-shot | 45.81±0.78 | 39.12±0.69 | 45.71±0.87 | 35.24±0.69 | 40.50±0.34 | 37.27±0.41 | 45.77±0.73 | 64.53±0.10 | 45.54±0.74 | 73.42±0.95 |
| 3-shot | 35.68±3.97 | 32.92±4.99 | 35.82±3.89 | 20.12±4.07 | 30.70±2.24 | 35.52±2.08 | 35.85±4.15 | 43.46±2.00 | 35.90±4.11 | 42.87±4.30 |
| *text-davinci-001* | | | | | | | | | | |
| 0-shot | 12.71±5.81 | 12.75±2.83 | 12.68±5.77 | 10.50±3.93 | 15.93±4.26 | 14.54±2.22 | 12.86±5.85 | 17.18±5.71 | 12.81±5.87 | 20.93±8.03 |
| 1-shot | 23.46±3.10 | 22.58±4.43 | 23.50±3.25 | 17.31±0.43 | 25.61±2.04 | 21.45±2.66 | 23.49±3.14 | 31.72±0.39 | 23.49±3.01 | 38.48±0.83 |
| 3-shot | 26.56±1.93 | 26.46±1.27 | 26.48±2.11 | 16.50±0.43 | 25.06±1.56 | 23.81±1.18 | 26.57±2.01 | 30.24±1.99 | 26.75±2.06 | 28.47±2.60 |
| *fine-tuned* | | | | | | | | | | |
| BERT-base-cased | 87.27 | 86.23 | 87.35 | **48.08** | 87.53 | **83.08** | 87.35 | 78.95 | 87.35 | **82.12** |
| BERT-base-uncased | **88.75** | **88.65** | 88.68 | 46.01 | **89.14** | 82.95 | 88.68 | 74.56 | **88.68** | 78.54 |
| ELMo-BiLSTM-CRF | 86.68 | 86.38 | 86.06 | 44.21 | 86.96 | 82.24 | 86.06 | 84.71 | 86.06 | 75.50 |
| Flair Embeddings | 85.53 | 85.21 | 84.57 | 44.94 | 86.07 | 81.54 | 84.57 | 81.26 | 84.57 | 73.14 |
| CNN-BiLSTM-CRF | 82.10 | 81.24 | 82.85 | 43.5 | 82.46 | 73.52 | 82.85 | 64.20 | 82.85 | 67.74 |
| GRN | 83.68 | 83.30 | 83.97 | 39.04 | 84.15 | 76.86 | 83.97 | **79.61** | 83.97 | 70.89 |

Table 16: Robustness test results of GPT series models (zero-shot and few-shot) and fine-tuned models (i.e., BERT-base-cased, BERT-base-uncased, ELMo-BiLSTM-CRF, Flair Embeddings, CNN-BiLSTM-CRF, GRN) on **CoNLL2003** dataset.

| Model | ConcatSent #3453 samples | | CrossCatagory #3453 samples | | EntTypos #2676 samples | | OOV #3453 samples | | SwapLonger #3453 samples | |
|---|---|---|---|---|---|---|---|---|---|---|
| | ori | trans | ori | trans | ori | trans | ori | trans | ori | trans |
| *text-davinci-003* | | | | | | | | | | |
| 0-shot | 50.46±2.10 | 52.97±1.27 | 50.47±2.15 | 26.39±1.50 | 55.93±1.81 | 43.43±1.31 | 50.42±2.21 | 47.13±1.55 | 50.43±2.14 | 43.29±1.32 |
| 1-shot | 52.40±1.24 | 53.96±1.98 | 52.35±1.29 | 27.11±1.72 | 57.54±1.14 | 47.84±1.11 | 52.39±1.11 | 47.95±1.25 | 52.36±1.13 | 46.13±1.77 |
| 3-shot | 57.73±1.50 | 56.80±1.58 | 57.62±1.58 | 30.93±0.38 | 61.13±1.55 | 49.65±1.40 | 57.70±1.56 | 56.21±1.86 | 57.70±1.60 | 52.34±0.32 |
| *text-davinci-002* | | | | | | | | | | |
| 0-shot | 54.69±1.28 | 58.04±0.93 | 54.91±1.41 | 29.19±0.52 | 57.04±0.75 | 49.04±0.66 | 54.79±1.37 | 55.25±0.51 | 54.79±1.28 | 52.14±0.61 |
| 1-shot | 56.70±1.50 | 55.76±2.09 | 56.73±1.10 | **35.40±1.22** | 58.70±1.71 | 51.53±2.57 | 56.69±1.44 | 58.46±1.97 | 56.82±1.36 | 59.20±0.82 |
| 3-shot | **61.64±0.87** | 59.15±0.84 | 61.52±0.94 | 33.21±0.15 | 63.20±0.93 | 54.58±1.83 | 61.50±0.84 | 64.52±0.27 | 61.52±0.92 | 58.78±0.47 |
| *text-davinci-001* | | | | | | | | | | |
| 0-shot | 18.92±1.12 | 18.48±1.64 | 18.84±1.48 | 11.94±0.45 | 22.00±0.97 | 18.61±1.24 | 18.72±0.95 | 13.83±1.35 | 19.12±1.17 | 16.36±1.07 |
| 1-shot | 28.66±1.42 | 26.06±2.43 | 28.07±1.50 | 19.39±0.85 | 30.85±1.38 | 27.80±1.69 | 28.75±1.44 | 28.04±2.16 | 28.44±1.17 | 33.17±2.45 |
| 3-shot | 35.69±0.35 | 30.10±1.73 | 35.53±0.60 | 19.32±0.76 | 37.50±0.47 | 29.81±1.57 | 35.54±0.27 | 28.60±1.81 | 35.71±0.67 | 39.36±0.16 |
| *fine-tuned* | | | | | | | | | | |
| BERT-base-cased | 91.43 | 89.91 | 91.42 | 44.42 | 92.20 | 85.02 | 91.42 | 68.71 | 91.42 | **79.28** |
| BERT-base-uncased | 90.41 | 90.05 | 90.40 | **47.19** | 91.25 | 81.25 | 90.40 | 64.46 | 90.40 | 78.26 |
| ELMo-BiLSTM-CRF | 91.80 | 90.67 | 91.79 | 44.13 | 92.48 | 86.19 | 91.79 | 68.10 | 91.79 | 61.82 |
| Flair Embeddings | **92.25** | **90.73** | 92.24 | 45.30 | **93.05** | **86.78** | 92.24 | 73.45 | 92.24 | 66.13 |
| CNN-BiLSTM-CRF | 90.61 | 87.99 | 90.59 | 44.18 | 91.25 | 79.10 | 90.59 | 58.99 | 90.59 | 61.15 |
| GRN | 91.57 | 89.30 | 91.56 | 42.90 | 92.29 | 82.72 | 91.56 | 68.20 | 91.56 | 65.38 |

Table 17: Robustness test results of GPT series models (zero-shot and few-shot) and fine-tuned models (i.e., BERT-base-cased, BERT-base-uncased, ELMo-BiLSTM-CRF, Flair Embeddings, CNN-BiLSTM-CRF, GRN) on **OntoNotesv5** dataset.

| Model | ConcatSent # 4019 samples | | CrossCatagory # 4019 samples | | EntTypos # 4492 samples | | OOV # 4019 samples | | SwapLonger # 4019 samples | |
|---|---|---|---|---|---|---|---|---|---|---|
| | ori | trans | ori | trans | ori | trans | ori | trans | ori | trans |
| *text-davinci-003* | | | | | | | | | | |
| 0-shot | 7.19±0.58 | 7.76±0.73 | 7.11±0.56 | 3.30±0.25 | 31.1±2.69 | 25.32±2.22 | 7.13±0.50 | 13.05±0.19 | 7.16±0.46 | 12.05±0.98 |
| 1-shot | 13.54±1.29 | 15.04±1.53 | 13.53±1.37 | 6.32±0.41 | **37.40±1.49** | **31.08±1.44** | 13.53±1.35 | 18.76±1.26 | 13.57±1.32 | 17.75±1.38 |
| 3-shot | **16.63±1.85** | **19.07±1.02** | **16.61±1.77** | **7.85±1.04** | 36.85±1.87 | 31.04±2.01 | **16.64±1.76** | **24.47±2.11** | **16.68±1.85** | **23.13±1.99** |
| *text-davinci-002* | | | | | | | | | | |
| 0-shot | 1.94±0.25 | 2.17±0.45 | 2.33±0.64 | 1.05±0.33 | 29.78±0.82 | 25.82±0.67 | 1.98±0.20 | 3.95±0.91 | 1.97±0.31 | 2.87±0.45 |
| 1-shot | 4.22±0.33 | 5.06±0.16 | 4.09±0.11 | 2.44±0.39 | 29.29±3.65 | 27.95±2.47 | 4.07±0.29 | 8.93±0.60 | 4.22±0.38 | 7.01±0.64 |
| 3-shot | 8.00±3.11 | 9.78±1.14 | 8.59±2.19 | 5.43±1.10 | 33.50±7.37 | 26.68±5.88 | 8.57±2.17 | 17.96±2.36 | 8.46±2.77 | 14.9±2.97 |
| *text-davinci-001* | | | | | | | | | | |
| 0-shot | 0.17±0.12 | 0.21±0.04 | 0.17±0.12 | 0.22±0.14 | 15.14±1.98 | 12.18±0.79 | 0.17±0.12 | 0.78±0.07 | 0.15±0.12 | 0.37±0.07 |
| 1-shot | 0.87±0.24 | 1.24±0.24 | 0.80±0.23 | 0.50±0.15 | 10.57±0.59 | 8.26±0.92 | 0.83±0.25 | 1.47±0.09 | 0.81±0.22 | 1.20±0.23 |
| 3-shot | 2.25±1.61 | 2.23±1.38 | 2.25±1.61 | 2.53±1.88 | 16.70±3.56 | 13.60±3.21 | 2.31±1.71 | 6.16±4.56 | 2.23±1.58 | 4.55±3.28 |
| *fine-tuned* | | | | | | | | | | |
| BERT-base-cased | **88.85** | **88.31** | **80.11** | **34.96** | **89.46** | 78.31 | **80.11** | **66.39** | **80.11** | **60.83** |
| BERT-base-uncased | 86.85 | 86.68 | 75.21 | 33.47 | 87.45 | 75.11 | 75.21 | 59.56 | 75.21 | 55.09 |
| ELMo-BiLSTM-CRF | 86.94 | 86.59 | 72.68 | 27.37 | 87.55 | 78.50 | 72.68 | 60.04 | 72.68 | 57.17 |
| Flair Embeddings | 86.96 | 86.67 | 70.65 | 25.78 | 87.66 | **80.12** | 70.65 | 55.76 | 70.65 | 55.12 |
| CNN-BiLSTM-CRF | 85.95 | 85.45 | 69.10 | 26.49 | 86.65 | 73.76 | 69.10 | 45.62 | 69.10 | 54.43 |
| GRN | 87.25 | 86.80 | 70.27 | 24.86 | 87.94 | 75.61 | 70.27 | 54.35 | 70.27 | 52.70 |

We evaluate the performance and robustness of GPT series models on six NER datasets, as shown in Table 1. In this subsection, we present the experimental results obtained from the first three NER datasets: ACE2005, CoNLL2003 and OntoNotesv5. The results of the last three NER datasets will be introduced in Section4.7.1. We use the TextFlint tool to perform five different transformations on the text for these datasets. All experimental results are given as micro-F1 scores. For details of our results, please refer to Table 15 ∼ 17.

Analyzing the experimental results, we can draw the following conclusions: **GPT series models face significant challenges in NER tasks.** GPT series models have much lower F1 scores compared to existing fine-tuned models. For example, the best scores for the GPT series models hover around 50 on the ACE2005 dataset (Table 15), while the fine-tuned models have achieved scores of nearly 90. However, increasing the number of examples in prompts can significantly improves the models' performance. This demonstrates that increasing the number of examples in the prompt has a positive effect on performance. **The performance of text-davinci-003 model is not always the best.** When comparing GPT models, for ACE2005 dataset (Table 15) and OntoNotesv5 dataset (Table 17), text-davinci-003 outperforms text-davinci-002 and text-davinci-001, indicating that as the models evolve from generation to generation, their ability to understand these NER tasks is increasing. However, for CoNLL2003 dataset (Table 16), text-davinci-002 performs the best among the three models in the GPT series. We speculate that in some NER tasks, the overall ability of the model may not be proportional to its performance on certain corpora. **The robustness of the GPT series models for NER tasks is unstable.** The various methods of text transformations used for different datasets in NER tasks demonstrate varying degrees of impact on robustness. In certain types of data transformations, there is a slight increase in robustness, while in others, there may be a slight decline or no significant change in robustness.

### 4.5.2 Part-of-speech Tagging

Table 18: Robustness test results of GPT series models (zero-shot and few-shot) and fine-tuned models (i.e., ELMo-BiLSTM-CRF, Flair-BiLSTM-CRF, CNN-BiLSTM-CRF, BERT-BiLSTM-CRF (Devlin et al., 2019)) on **WSJ** dataset. The model tagged with † means that the example in the prompt uses the original labels (e.g. NN, etc.).Since this task does not have a specific transformation for the full amount of data, we test the performance of the model on it separately (Column all). The missing text-davinci-001 results are due to the fact that the model does not do the task according to the prompt.

(a)

| Model | SwapMultiPOSJJ # 3963 samples | | SwapMultiPOSNN # 4952 samples | | SwapMultiPOSRB # 2874 samples | |
|---|---|---|---|---|---|---|
| | ori | trans | ori | trans | ori | trans |
| *text-davinci-003* | | | | | | |
| 0-shot | 75.67±2.65 | 74.92±2.79 | 75.47±2.63 | 74.26±2.72 | 74.81±2.60 | 72.80±2.81 |
| 1-shot | 71.42±0.69 | 70.71±0.67 | 71.35±0.64 | 70.46±0.71 | 71.40±0.55 | 69.87±0.5 4 |
| 3-shot | 84.09±0.21 | 83.57±0.17 | 83.91±0.20 | 83.07±0.16 | 83.55±0.29 | 81.86±0.33 |
| 1-shot† | **88.77±0.62** | **88.31±0.57** | **88.18±0.57** | **87.21±0.53** | **87.64±0.48** | **85.69±0.31** |
| *text-davinci-002* | | | | | | |
| 0-shot | 71.48±0.56 | 70.78±0.69 | 71.13±0.67 | 70.06±0.76 | 70.62±0.52 | 69.14±0.77 |
| 1-shot | 68.75±0.71 | 67.68±0.55 | 68.29±0.50 | 67.20±0.38 | 68.06±0.51 | 66.26±0.37 |
| 3-shot | 79.64±1.00 | 79.28±1.11 | 79.53±1.06 | 78.85±0.95 | 79.74±1.00 | 77.99±0.66 |
| *text-davinci-001* | | | | | | |
| 0/1/3-shot | - | - | - | - | - | - |
| *fine-tuned* | | | | | | |
| ELMo-BiLSTM-CRF | 97.73 | **96.92** | 97.72 | 96.48 | 97.55 | 95.11 |
| Flair-BiLSTM-CRF | 97.73 | 96.58 | **97.74** | **97.08** | **97.58** | 95.03 |
| CNN-BiLSTM-CRF | 97.55 | 96.46 | 97.53 | 96.90 | 97.35 | 94.51 |
| BERT-BiLSTM-CRF | **97.78** | **96.92** | 97.72 | 96.48 | 97.57 | **97.75** |

(b)

| Model | SwapMultiPOSVB # 2376 samples | | SwapPrefix # 4526 samples | | all # 5461 samples |
|---|---|---|---|---|---|
| | ori | trans | ori | trans | ori |
| *text-davinci-003* | | | | | |
| 0-shot | 76.21±2.61 | 76.21±2.61 | 75.51±2.73 | 74.94±2.74 | 75.02±2.59 |
| 1-shot | 72.88±0.69 | 72.88±0.69 | 71.46±0.63 | 71.08±0.64 | 70.79±0.71 |
| 3-shot | 84.61±0.27 | 84.61±0.27 | 84.07±0.20 | 83.67±0.19 | **83.69±0.16** |
| 1-shot† | **87.75±0.78** | **87.70±0.55** | **88.40±0.63** | **88.10±0.57** | - |
| *text-davinci-002* | | | | | |
| 0-shot | 70.73±0.81 | 70.73±0.81 | 71.43±0.77 | 70.66±0.77 | 71.02±0.62 |
| 1-shot | 68.91±0.47 | 68.91±0.47 | 68.70±0.34 | 68.21±0.50 | 68.13±0.46 |
| 3-shot | 81.09±0.92 | 81.09±0.92 | 80.10±1.06 | 79.73±1.02 | 79.48±1.11 |
| *text-davinci-001* | | | | | |
| 0/1/3-shot | - | - | - | - | - |
| *fine-tuned* | | | | | |
| CNN-BiLSTM-CRF | 97.63 | 97.30 | 97.55 | 96.87 | 97.54 |
| ELMo-BiLSTM-CRF | 97.80 | 97.39 | 97.75 | 96.66 | **97.75** |
| Flair-BiLSTM-CRF | **97.83** | **97.53** | **97.76** | **97.33** | **97.75** |
| BERT-BiLSTM-CRF | 97.80 | 93.34 | 97.75 | 96.66 | **97.75** |

We evaluate the performance and robustness of GPT series models using three POS datasets including two English datasets (i.e., WSJ and Daily547) and one Chinese dataset (i.e., PKU-SEGPOS). We analyze the results of the WSJ dataset in detail and provide more detailed results on Daily547 and PKU-SEGPOS in Section 4.7.1. We also perform five text transformations on the WSJ dataset using the TextFlint tool to evaluate the robustness of the models. All experimental results are given as accuracy and are presented in Table 18. Our analysis of the results reveals the following conclusions. **First, the GPT series models also face challenges in POS tasks.** The accuracy for various forms of text transformation on the WSJ dataset using the GPT series of models remain inferior to those of existing fine-tuned models, but the gap has significantly decreased compared to the NER tasks. The 1-shot model shows decreased performance compared to the 0-shot model, while the 3-shot model showes a significant improvement in performance. **Second, models require sufficient semantic understanding to perform complex tasks like POS.** Comparing GPT models, there is only a slight performance improvement in text-davinci-003 compared to text-davinci-002, which both belong to GPT-3.5 series. However, the GPT-3 series model, text-davinci-001, is unable to complete the task, as it provides an excessive number of non-standard format answers beyond the threshold. [9] In other tasks, there has never been the phenomenon where the model completely fails to understand all of the zero/few-shot prompts. This also indicates that it is not until the text-davinci-002 generation that the GPT models achieved an understanding and basic completion of the corresponding POS tasks. We speculate that this may be because the model has been trained on code since text-davinci-002 and further research is needed to confirm this. **Third, the GPT-3.5 models show good robustness in POS tasks.** The text-davinci-003 and text-davinci-002 models in the GPT-3.5 series, as well as the fine-tuned models, have demonstrated strong robustness against various forms of text transformations in this task, with their robustness being roughly equal.

## 4.6 The Winograd Schema Challenge

Table 19: Robustness test results of GPT series models (zero-shot and few-shot) and fine-tuned models (i.e., BERT-base-uncased, BERT-large-uncased, HNN (He et al., 2019)) on **WSC273** dataset.

| Model | all<br># 570 samples | AddSentences<br># 570 samples | InsertRelativeClause<br># 566 samples | SwapNames<br># 566 samples | SwitchVoice<br># 440 samples | SwapGender<br># 310 samples |
|---|---|---|---|---|---|---|
| | | | *text-davinci-003* | | | |
| 0-shot | 62.05±0.57 | 65.32±1.77 | **59.83±0.51** | 60.48±0.74 | 59.39±0.92 | 36.99±1.30 |
| 1-shot | 61.40±1.37 | 62.98±0.63 | 58.18±0.57 | 58.42±0.44 | 57.35±0.13 | 38.28±0.49 |
| 3-shot | **62.75±1.13** | **65.38±0.53** | 57.95±0.47 | 61.37±0.80 | **59.62±0.57** | 36.77±0.65 |
| | | | *text-davinci-002* | | | |
| 0-shot | 61.46±1.57 | 64.09±1.00 | 56.89±1.23 | 59.84±1.77 | 59.47±1.77 | 39.03±2.01 |
| 1-shot | 60.94±2.03 | 65.20±3.62 | 58.54±2.58 | 59.54±2.46 | 57.80±1.26 | 38.39±1.48 |
| 3-shot | 62.22±1.47 | 64.91±0.88 | 59.07±2.12 | **62.07±0.89** | 58.18±1.42 | 38.93±2.44 |
| | | | *text-davinci-001* | | | |
| 0-shot | 52.05±1.14 | 53.22±1.60 | 50.94±1.37 | 51.12±1.41 | 51.14±0.82 | 47.96±0.38 |
| 1-shot | 50.41±0.56 | 53.39±2.78 | 50.18±0.36 | 49.76±0.10 | 50.23±0.39 | 49.35±0.65 |
| 3-shot | 51.34±0.97 | 54.79±2.11 | 51.47±0.84 | 51.65±0.80 | 50.83±1.84 | **49.57±1.83** |
| | | | *fine-tuned* | | | |
| BERT-base-uncased | 56.00 | - | 57.60 | 52.70 | 57.70 | **46.45** |
| BERT-large-uncased | 61.90 | - | 59.10 | 56.50 | 60.90 | 41.94 |
| HNN | **75.10** | - | **67.50** | **69.60** | **67.70** | 27.10 |

For WSC, we do experiments on WSC273 dataset and report the accuracy in Table 19. Analyzing the results, we have the following findings. **The performance of text-davinci-003 can be compared with BERT-large on the WSC task.** Although the performance of text-davinci-003 on WSC273 is still significantly different from the SOTA results, it is close to or exceeds the BERT-large results, indicating some potential in tackling the challenge. **However, the enhanced capability of the GPT models leads to reduced robustness in some text transformations.** The model capability of text-davinci-001 to text-davinci-003 is progressively enhanced, and it is true that text-davinci-003 achieves the best performance on the original dataset. However, for the datasets related to name and gender, text-davinci-002 and text-davinci-001 achieve the best results, respectively. This suggests that the enhanced capability of our model has lost some careful judgment and needs to be further investigated.

---
[9]When the GPT model exhibits this behavior, we use the symbol '-' in the table to represent it, the same as below.

### 4.7 Further Studies

#### 4.7.1 Partial Sequence Tagging Task

Table 20: Performance test results of GPT series models (zero-shot and few-shot) on partial sequence tagging datasets, using accuracy as the evaluation for Daily547 and micro-F1 score for others. Some of the missing results in the table are due to the fact that the model does not complete the task according to the prompt resulting in our inability to analyze the results.

| Model | HONOR<br>*# 1120 samples* | MSRANER<br>*# 4365 samples* | OntoNote4NER<br>*# 4346 samples* | Daily547<br>*# 546 samples* | PKU-SEGPOS<br>*# 5204 samples* |
|---|---|---|---|---|---|
| | | | *text-davinci-003* | | |
| 0-shot | 47.62±2.00 | 23.02±4.67 | 30.66±1.19 | 64.80±0.18 | 65.86±1.18 |
| 1-shot | 49.04±0.89 | 46.89±0.81 | 49.72±0.81 | 77.99±1.15 | 76.43±0.45 |
| 3-shot | **54.01±2.22** | 57.14±0.36 | **50.98±0.36** | **82.63±0.55** | **76.88±0.36** |
| | | | *text-davinci-002* | | |
| 0-shot | 46.60±8.34 | 15.13±1.83 | - | 52.96±4.49 | 39.11±6.12 |
| 1-shot | 45.14±11.55 | 35.63±2.14 | 35.69±4.34 | 65.25±0.62 | 56.28±1.55 |
| 3-shot | 51.02±13.63 | **58.24±1.88** | 34.27±11.17 | 76.03±0.51 | 51.82±2.00 |
| | | | *text-davinci-001* | | |
| 0-shot | - | - | - | - | - |
| 1-shot | 32.42±1.05 | 25.16±2.18 | - | - | 15.09±0.37 |
| 3-shot | 39.09±0.40 | 42.43±0.68 | 36.38±0.23 | - | 24.30±0.81 |

We test the performance of GPT series of models on few-shot scenarios of a range of NER and POS datasets and obtain Table 20. **The results show that their scores remain quite low, which further confirms our conclusion that the GPT-3.5 model faces considerable difficulties in the ST task (and the GPT-3 model, even more so)**. Nevertheless, increasing the number of examples in prompts leads to a significant improvement in the scores for most of datasets. The text-davinci-003 model achieves the best performance overall in the 3-shot scenario compared to its own performance in all these tasks.

#### 4.7.2 Comparison of text-davinci-003 and other LLMs

Table 21: Zero-shot and few-shot performance of text-davinci-003, LaMDA-PT (Thoppilan et al., 2022) and FLAN.

| Subtask | Dataset | text-davinci-003 | | LaMDA-PT | | FLAN | |
|---|---|---|---|---|---|---|---|
| | | 0-shot | 3-shot | 0-shot | [k]-shot | 0-shot | [k]-shot |
| Machine Reading Comprehension | SQuAD1.1 | 78.22 | **89.80** | 22.70 | 50.20 [3] | 80.00 | 82.70 [4] |
| | SQuAD2.0 | 76.46 | **89.21** | 11.10 | 34.90 [3] | 44.20 | 43.10 [3] |
| Natural Language Inference | MNLI-m | 64.67 | **78.62** | 35.70 | 43.70 [5] | 61.20 | 63.50 [10] |
| | MNLI-mm | 64.93 | **77.51** | 37.00 | 43.80 [5] | 62.40 | 63.50 [10] |
| | SNLI | 69.49 | **74.73** | 33.30 | 54.70 [5] | 53.40 | 65.60 [15] |
| Semantic Matching | MRPC | 74.50 | **79.00** | 53.70 | 64.00 [5] | 69.10 | 67.20 [10] |
| | QQP | 81.60 | **84.20** | 34.90 | 58.90 [3] | 75.90 | 75.90 [16] |
| Sentiment Classification | IMDB | 92.50 | 88.69 | 76.90 | 83.30 [1] | 94.30 | **95.00** [2] |

We compare the performance of text-davinci-003 with that of other LLMs (i.e., LaMDA-PT and FLAN) on some tasks and present the results in Table 21. From the table, we can observe that text-davinci-003 outperforms the best performance of the original LLMs in few-shot scenarios on most tasks at 3-shot, and even surpasses the best performance of the original LLMs at 0-shot on some tasks (e.g., NLI and SM), demonstrating its superior understanding compared to the original LLMs.

## 5 Conclusion

This study provides a comprehensive and in-depth analysis of the robustness of GPT-3.5 models. Our investigation covers 21 datasets across 9 different NLU tasks, providing a broad understanding of the models' robustness. Our results show that while GPT-3.5 performs better than existing fine-tuned models in certain tasks, it still faces significant robustness degradation and specific robustness challenges, including robustness instability, prompt sensitivity, number sensitivity, and sensitivity to task labels and types. We hope that our study will inspire future research to 1) examine additional aspects of robustness, such as the robustness of instructions; 2) explore more effective prompting strategies that ensure consistency between the instruction tuning stage and the actual application stage of the model; and 3) design better prompts for certain tasks, such as NER and RE, taking into account the complexity of the instruction construction which may be a contributing factor to their lower performance.

## References

Alan Akbik, Duncan Blythe, and Roland Vollgraf. 2018. Contextual string embeddings for sequence labeling. In *Proceedings of the 27th international conference on computational linguistics*, pages 1638–1649.

Yejin Bang, Samuel Cahyawijaya, Nayeon Lee, Wenliang Dai, Dan Su, Bryan Wilie, Holy Lovenia, Ziwei Ji, Tiezheng Yu, Willy Chung, Quyet V. Do, Yan Xu, and Pascale Fung. 2023. A multitask, multilingual, multimodal evaluation of chatgpt on reasoning, hallucination, and interactivity. *CoRR*, abs/2302.04023.

Tom B. Brown, Benjamin Mann, Nick Ryder, Melanie Subbiah, Jared Kaplan, Prafulla Dhariwal, Arvind Neelakantan, Pranav Shyam, Girish Sastry, Amanda Askell, Sandhini Agarwal, Ariel Herbert-Voss, Gretchen Krueger, Tom Henighan, Rewon Child, Aditya Ramesh, Daniel M. Ziegler, Jeffrey Wu, Clemens Winter, Christopher Hesse, Mark Chen, Eric Sigler, Mateusz Litwin, Scott Gray, Benjamin Chess, Jack Clark, Christopher Berner, Sam McCandlish, Alec Radford, Ilya Sutskever, and Dario Amodei. 2020. Language models are few-shot learners. In Hugo Larochelle, Marc'Aurelio Ranzato, Raia Hadsell, Maria-Florina Balcan, and Hsuan-Tien Lin, editors, *Advances in Neural Information Processing Systems 33: Annual Conference on Neural Information Processing Systems 2020, NeurIPS 2020, December 6-12, 2020, virtual*.

Hui Chen, Zijia Lin, Guiguang Ding, Jianguang Lou, Yusen Zhang, and Borje Karlsson. 2019. Grn: Gated relation network to enhance convolutional neural network for named entity recognition. In *Proceedings of the AAAI Conference on Artificial Intelligence*, volume 33, pages 6236–6243.

Aakanksha Chowdhery, Sharan Narang, Jacob Devlin, Maarten Bosma, Gaurav Mishra, Adam Roberts, Paul Barham, Hyung Won Chung, Charles Sutton, Sebastian Gehrmann, Parker Schuh, Kensen Shi, Sasha Tsvyashchenko, Joshua Maynez, Abhishek Rao, Parker Barnes, Yi Tay, Noam Shazeer, Vinodkumar Prabhakaran, Emily Reif, Nan Du, Ben Hutchinson, Reiner Pope, James Bradbury, Jacob Austin, Michael Isard, Guy Gur-Ari, Pengcheng Yin, Toju Duke, Anselm Levskaya, Sanjay Ghemawat, Sunipa Dev, Henryk Michalewski, Xavier Garcia, Vedant Misra, Kevin Robinson, Liam Fedus, Denny Zhou, Daphne Ippolito, David Luan, Hyeontaek Lim, Barret Zoph, Alexander Spiridonov, Ryan Sepassi, David Dohan, Shivani Agrawal, Mark Omernick, Andrew M. Dai, Thanumalayan Sankaranarayana Pillai, Marie Pellat, Aitor Lewkowycz, Erica Moreira, Rewon Child, Oleksandr Polozov, Katherine Lee, Zongwei Zhou, Xuezhi Wang, Brennan Saeta, Mark Diaz, Orhan Firat, Michele Catasta, Jason Wei, Kathy Meier-Hellstern, Douglas Eck, Jeff Dean, Slav Petrov, and Noah Fiedel. 2022. Palm: Scaling language modeling with pathways. *CoRR*, abs/2204.02311.

Paul F. Christiano, Jan Leike, Tom B. Brown, Miljan Martic, Shane Legg, and Dario Amodei. 2017. Deep reinforcement learning from human preferences. In Isabelle Guyon, Ulrike von Luxburg, Samy Bengio, Hanna M. Wallach, Rob Fergus, S. V. N. Vishwanathan, and Roman Garnett, editors, *Advances in Neural Information Processing Systems 30: Annual Conference on Neural Information Processing Systems 2017, December 4-9, 2017, Long Beach, CA, USA*, pages 4299–4307.

Kevin Clark, Minh-Thang Luong, Quoc V. Le, and Christopher D. Manning. 2020. ELECTRA: pre-training text encoders as discriminators rather than generators. *CoRR*, abs/2003.10555.

Zihang Dai, Guokun Lai, Yiming Yang, and Quoc V. Le. 2020. Funnel-transformer: Filtering out sequential redundancy for efficient language processing. *CoRR*, abs/2006.03236.

Jacob Devlin, Ming-Wei Chang, Kenton Lee, and Kristina Toutanova. 2019. BERT: pre-training of deep bidirectional transformers for language understanding. In Jill Burstein, Christy Doran, and Thamar Solorio, editors, *Proceedings of the 2019 Conference of the North American Chapter of the Association for Computational Linguistics: Human Language Technologies, NAACL-HLT 2019, Minneapolis, MN, USA, June 2-7, 2019, Volume 1 (Long and Short Papers)*, pages 4171–4186. Association for Computational Linguistics.

Bill Dolan and Chris Brockett. 2005. Automatically constructing a corpus of sentential paraphrases. In *Third International Workshop on Paraphrasing (IWP2005)*.

Javid Ebrahimi, Anyi Rao, Daniel Lowd, and Dejing Dou. 2018. Hotflip: White-box adversarial examples for text classification. In Iryna Gurevych and Yusuke Miyao, editors, *Proceedings of the 56th Annual Meeting of the Association for Computational Linguistics, ACL 2018, Melbourne, Australia, July 15-20, 2018, Volume 2: Short Papers*, pages 31–36. Association for Computational Linguistics.

Nanyi Fei, Zhiwu Lu, Yizhao Gao, Guoxing Yang, Yuqi Huo, Jingyuan Wen, Haoyu Lu, Ruihua Song, Xin Gao, Tao Xiang, et al. 2022. Towards artificial general intelligence via a multimodal foundation model. *Nature Communications*, 13(1):3094.

Simon Frieder, Luca Pinchetti, Ryan-Rhys Griffiths, Tommaso Salvatori, Thomas Lukasiewicz, Philipp Christian Petersen, Alexis Chevalier, and Julius Berner. 2023. Mathematical capabilities of chatgpt. *arXiv preprint arXiv:2301.13867*.

Kevin Gimpel, Nathan Schneider, Brendan O'Connor, Dipanjan Das, Daniel Mills, Jacob Eisenstein, Michael Heilman, Dani Yogatama, Jeffrey Flanigan, and Noah A Smith. 2010. Part-of-speech tagging for twitter: Annotation, features, and experiments. Technical report, Carnegie-Mellon Univ Pittsburgh Pa School of Computer Science.

Ben Goertzel. 2014. Artificial general intelligence: concept, state of the art, and future prospects. *Journal of Artificial General Intelligence*, 5(1):1.

Tao Gui, Xiao Wang, Qi Zhang, Qin Liu, Yicheng Zou, Xin Zhou, Rui Zheng, Chong Zhang, Qinzhuo Wu, Jiacheng Ye, et al. 2021. Textflint: Unified multilingual robustness evaluation toolkit for natural language processing. *arXiv preprint arXiv:2103.11441*.

Biyang Guo, Xin Zhang, Ziyuan Wang, Minqi Jiang, Jinran Nie, Yuxuan Ding, Jianwei Yue, and Yupeng Wu. 2023. How close is chatgpt to human experts? comparison corpus, evaluation, and detection.

Pengcheng He, Xiaodong Liu, Weizhu Chen, and Jianfeng Gao. 2019. A hybrid neural network model for commonsense reasoning. *CoRR*, abs/1907.11983.

Pengcheng He, Xiaodong Liu, Jianfeng Gao, and Weizhu Chen. 2020. Deberta: Decoding-enhanced BERT with disentangled attention. *CoRR*, abs/2006.03654.

Amr Hendy, Mohamed Abdelrehim, Amr Sharaf, Vikas Raunak, Mohamed Gabr, Hitokazu Matsushita, Young Jin Kim, Mohamed Afify, and Hany Hassan Awadalla. 2023. How good are gpt models at machine translation? a comprehensive evaluation. *arXiv preprint arXiv:2302.09210*.

Jeremy Howard and Sebastian Ruder. 2018. Fine-tuned language models for text classification. *CoRR*, abs/1801.06146.

Di Jin, Zhijing Jin, Joey Tianyi Zhou, and Peter Szolovits. 2020. Is BERT really robust? A strong baseline for natural language attack on text classification and entailment. In *The Thirty-Fourth AAAI Conference on Artificial Intelligence, AAAI 2020, The Thirty-Second Innovative Applications of Artificial Intelligence Conference, IAAI 2020, The Tenth AAAI Symposium on Educational Advances in Artificial Intelligence, EAAI 2020, New York, NY, USA, February 7-12, 2020*, pages 8018–8025. AAAI Press.

Mandar Joshi, Danqi Chen, Yinhan Liu, Daniel S. Weld, Luke Zettlemoyer, and Omer Levy. 2020. Spanbert: Improving pre-training by representing and predicting spans. *Trans. Assoc. Comput. Linguistics*, 8:64–77.

Jared Kaplan, Sam McCandlish, Tom Henighan, Tom B. Brown, Benjamin Chess, Rewon Child, Scott Gray, Alec Radford, Jeffrey Wu, and Dario Amodei. 2020. Scaling laws for neural language models. *CoRR*, abs/2001.08361.

Michal Kosinski. 2023. Theory of mind may have spontaneously emerged in large language models. *CoRR*, abs/2302.02083.

Zhenzhong Lan, Mingda Chen, Sebastian Goodman, Kevin Gimpel, Piyush Sharma, and Radu Soricut. 2019. ALBERT: A lite BERT for self-supervised learning of language representations. *CoRR*, abs/1909.11942.

Hector Levesque, Ernest Davis, and Leora Morgenstern. 2012. The winograd schema challenge. In *Thirteenth International Conference on the Principles of Knowledge Representation and Reasoning*. Citeseer.

Gina-Anne Levow. 2006. The third international chinese language processing bakeoff: Word segmentation and named entity recognition. In *Proceedings of the Fifth SIGHAN workshop on Chinese language processing*, pages 108–117.

Yinhan Liu, Myle Ott, Naman Goyal, Jingfei Du, Mandar Joshi, Danqi Chen, Omer Levy, Mike Lewis, Luke Zettlemoyer, and Veselin Stoyanov. 2019. Roberta: A robustly optimized BERT pretraining approach. *CoRR*, abs/1907.11692.

Haochen Liu, Yiqi Wang, Wenqi Fan, Xiaorui Liu, Yaxin Li, Shaili Jain, Yunhao Liu, Anil Jain, and Jiliang Tang. 2022. Trustworthy ai: A computational perspective. *ACM Transactions on Intelligent Systems and Technology*, 14(1):1–59.

Xuezhe Ma and Eduard Hovy. 2016. End-to-end sequence labeling via bi-directional lstm-cnns-crf. *arXiv preprint arXiv:1603.01354*.

Andrew Maas, Raymond E Daly, Peter T Pham, Dan Huang, Andrew Y Ng, and Christopher Potts. 2011. Learning word vectors for sentiment analysis. In *Proceedings of the 49th annual meeting of the association for computational linguistics: Human language technologies*, pages 142–150.

Mitchell P. Marcus, Beatrice Santorini, and Mary Ann Marcinkiewicz. 1993. Building a large annotated corpus of English: The Penn Treebank. *Computational Linguistics*, 19(2):313–330.

Matthew E. Peters, Mark Neumann, Mohit Iyyer, Matt Gardner, Christopher Clark, Kenton Lee, and Luke Zettlemoyer. 2018. Deep contextualized word representations. In Marilyn A. Walker, Heng Ji, and Amanda Stent, editors, *Proceedings of the 2018 Conference of the North American Chapter of the Association for Computational Linguistics: Human Language Technologies, NAACL-HLT 2018, New Orleans, Louisiana, USA, June 1-6, 2018, Volume 1 (Long Papers)*, pages 2227–2237. Association for Computational Linguistics.

Maria Pontiki, Dimitris Galanis, John Pavlopoulos, Harris Papageorgiou, Ion Androutsopoulos, and Suresh Manandhar. 2014. SemEval-2014 task 4: Aspect based sentiment analysis. In *Proceedings of the 8th International Workshop on Semantic Evaluation (SemEval 2014)*, pages 27–35, Dublin, Ireland, August. Association for Computational Linguistics.

Christopher Potts, Zhengxuan Wu, Atticus Geiger, and Douwe Kiela. 2020. Dynasent: A dynamic benchmark for sentiment analysis. *arXiv preprint arXiv:2012.15349*.

Chengwei Qin, Aston Zhang, Zhuosheng Zhang, Jiaao Chen, Michihiro Yasunaga, and Diyi Yang. 2023. Is chatgpt a general-purpose natural language processing task solver? *arXiv preprint arXiv:2302.06476*.

Pranav Rajpurkar, Jian Zhang, Konstantin Lopyrev, and Percy Liang. 2016. Squad: 100, 000+ questions for machine comprehension of text. *CoRR*, abs/1606.05250.

Pranav Rajpurkar, Robin Jia, and Percy Liang. 2018. Know what you don't know: Unanswerable questions for squad. *CoRR*, abs/1806.03822.

Marco Tulio Ribeiro, Tongshuang Wu, Carlos Guestrin, and Sameer Singh. 2020. Beyond accuracy: Behavioral testing of nlp models with checklist. *arXiv preprint arXiv:2005.04118*.

Erik Tjong Kim Sang and Fien De Meulder. 2003. Introduction to the conll-2003 shared task: language-independent named entity recognition. *North American Chapter of the Association for Computational Linguistics*.

Victor Sanh, Lysandre Debut, Julien Chaumond, and Thomas Wolf. 2019. Distilbert, a distilled version of BERT: smaller, faster, cheaper and lighter. *CoRR*, abs/1910.01108.

Victor Sanh, Albert Webson, Colin Raffel, Stephen H. Bach, Lintang Sutawika, Zaid Alyafeai, Antoine Chaffin, Arnaud Stiegler, Teven Le Scao, Arun Raja, Manan Dey, M Saiful Bari, Canwen Xu, Urmish Thakker, Shanya Sharma, Eliza Szczechla, Taewoon Kim, Gunjan Chhablani, Nihal V. Nayak, Debajyoti Datta, Jonathan Chang, Mike Tian-Jian Jiang, Han Wang, Matteo Manica, Sheng Shen, Zheng Xin Yong, Harshit Pandey, Rachel Bawden, Tom Wang, Trishala Neeraj, Jos Rozen, Abheesht Sharma, Andrea Santilli, Thibault Févry, Jason A. Fries, Ryan Teehan, Stella Biderman, Leo Gao, Tali Bers, Thomas Wolf, and Alexander M. Rush. 2021. Multitask prompted training enables zero-shot task generalization.

Youwei Song, Jiahai Wang, Tao Jiang, Zhiyue Liu, and Yanghui Rao. 2019. Attentional encoder network for targeted sentiment classification. *CoRR*, abs/1902.09314.

Chi Sun, Xipeng Qiu, Yige Xu, and Xuanjing Huang. 2019. How to fine-tune BERT for text classification? In Maosong Sun, Xuanjing Huang, Heng Ji, Zhiyuan Liu, and Yang Liu, editors, *Chinese Computational Linguistics - 18th China National Conference, CCL 2019, Kunming, China, October 18-20, 2019, Proceedings*, volume 11856 of *Lecture Notes in Computer Science*, pages 194–206. Springer.

Romal Thoppilan, Daniel De Freitas, Jamie Hall, Noam Shazeer, Apoorv Kulshreshtha, Heng-Tze Cheng, Alicia Jin, Taylor Bos, Leslie Baker, Yu Du, YaGuang Li, Hongrae Lee, Huaixiu Steven Zheng, Amin Ghafouri, Marcelo Menegali, Yanping Huang, Maxim Krikun, Dmitry Lepikhin, James Qin, Dehao Chen, Yuanzhong Xu, Zhifeng Chen, Adam Roberts, Maarten Bosma, Yanqi Zhou, Chung-Ching Chang, Igor Krivokon, Will Rusch, Marc Pickett, Kathleen S. Meier-Hellstern, Meredith Ringel Morris, Tulsee Doshi, Renelito Delos Santos, Toju Duke, Johnny Soraker, Ben Zevenbergen, Vinodkumar Prabhakaran, Mark Diaz, Ben Hutchinson, Kristen Olson, Alejandra Molina, Erin Hoffman-John, Josh Lee, Lora Aroyo, Ravi Rajakumar, Alena Butryna, Matthew Lamm, Viktoriya Kuzmina, Joe Fenton, Aaron Cohen, Rachel Bernstein, Ray Kurzweil, Blaise Aguera-Arcas, Claire Cui, Marian Croak, Ed H. Chi, and Quoc Le. 2022. Lamda: Language models for dialog applications. *CoRR*, abs/2201.08239.

Zhiguo Wang, Wael Hamza, and Radu Florian. 2017. Bilateral multi-perspective matching for natural language sentences. *CoRR*, abs/1702.03814.

Jason Wei, Maarten Bosma, Vincent Y. Zhao, Kelvin Guu, Adams Wei Yu, Brian Lester, Nan Du, Andrew M. Dai, and Quoc V. Le. 2022. Finetuned language models are zero-shot learners. In *The Tenth International Conference on Learning Representations, ICLR 2022, Virtual Event, April 25-29, 2022*. OpenReview.net.

Ralph Weischedel, Sameer Pradhan, Lance Ramshaw, Martha Palmer, Nianwen Xue, Mitchell Marcus, Ann Taylor, Craig Greenberg, Eduard Hovy, Robert Belvin, et al. 2011. Ontonotes release 4.0. *LDC2011T03, Philadelphia, Penn.: Linguistic Data Consortium*.

Adina Williams, Nikita Nangia, and Samuel R. Bowman. 2017. A broad-coverage challenge corpus for sentence understanding through inference. *CoRR*, abs/1704.05426.

Ikuya Yamada, Akari Asai, Hiroyuki Shindo, Hideaki Takeda, and Yuji Matsumoto. 2020. LUKE: deep contextualized entity representations with entity-aware self-attention. *CoRR*, abs/2010.01057.

Zhilin Yang, Zihang Dai, Yiming Yang, Jaime G. Carbonell, Ruslan Salakhutdinov, and Quoc V. Le. 2019. Xlnet: Generalized autoregressive pretraining for language understanding. *CoRR*, abs/1906.08237.

Yuhao Zhang, Victor Zhong, Danqi Chen, Gabor Angeli, and Christopher D. Manning. 2017. Position-aware attention and supervised data improve slot filling. In Martha Palmer, Rebecca Hwa, and Sebastian Riedel, editors, *Proceedings of the 2017 Conference on Empirical Methods in Natural Language Processing, EMNLP 2017, Copenhagen, Denmark, September 9-11, 2017*, pages 35–45. Association for Computational Linguistics.

Chen Zhang, Qiuchi Li, and Dawei Song. 2019. Aspect-based sentiment classification with aspect-specific graph convolutional networks. In *Proceedings of the 2019 Conference on Empirical Methods in Natural Language Processing and the 9th International Joint Conference on Natural Language Processing (EMNLP-IJCNLP)*, pages 4568–4578, Hong Kong, China, November. Association for Computational Linguistics.

Susan Zhang, Stephen Roller, Naman Goyal, Mikel Artetxe, Moya Chen, Shuohui Chen, Christopher Dewan, Mona T. Diab, Xian Li, Xi Victoria Lin, Todor Mihaylov, Myle Ott, Sam Shleifer, Kurt Shuster, Daniel Simig, Punit Singh Koura, Anjali Sridhar, Tianlu Wang, and Luke Zettlemoyer. 2022. OPT: open pre-trained transformer language models. *CoRR*, abs/2205.01068.

Wenxuan Zhou and Muhao Chen. 2021. Learning from noisy labels for entity-centric information extraction. In Marie-Francine Moens, Xuanjing Huang, Lucia Specia, and Scott Wen-tau Yih, editors, *Proceedings of the 2021 Conference on Empirical Methods in Natural Language Processing, EMNLP 2021, Virtual Event / Punta Cana, Dominican Republic, 7-11 November, 2021*, pages 5381–5392. Association for Computational Linguistics.

# A Details of Performance

Table 22 shows the best performance of the SOTA model, BERT, text-davinci-003 (0/3/6/9-shot) on each dataset, with its visualization results plotted in Figure 1.

Table 22: Details of performance. The result tagged with † is from BERT-Large-ITPT, and the result tagged with ‡ is from BERT-BILSTM-CRF.

| Task | Subtask | Dataset | SOTA | BERT | text-davinci-003 | | | |
| --- | --- | --- | --- | --- | --- | --- | --- | --- |
| | | | | | 0-shot | 3-shot | 6-shot | 9-shot |
| Machine Reading Comprehension | Machine Reading Comprehension | SQuAD1.1 | 90.07 | 87.09 | 78.22 | 89.80 | 91.43 | - |
| | | SQuAD2.0 | 84.62 | 75.95 | 76.46 | 89.21 | 79.23 | - |
| Relation Extraction | Relation Extraction | Tacred | 70.87 | 68.01 | 21.90 | 22.12 | 25.77 | 23.40 |
| Sentence Pair Relationship | Natural Language Inference | MNLI-m | 91.60 | 84.39 | 64.67 | 78.62 | 80.94 | 81.20 |
| | | MNLI-mm | 91.90 | 84.43 | 64.93 | 77.51 | 78.70 | 80.90 |
| | | SNLI | 90.60 | 88.99 | 69.49 | 74.73 | 81.00 | 82.20 |
| | Semantic Matching | QQP | 92.28 | 90.91 | 81.60 | 84.20 | 85.10 | 85.90 |
| | | MRPC | 88.40 | 84.28 | 74.50 | 79.00 | 77.60 | 78.20 |
| Sentiment Analysis | Aspect-based Sentiment Analysis | laptop | 91.79 | 83.69 | 84.12 | 86.05 | 86.70 | 86.91 |
| | | restaurant | 94.36 | 90.44 | 90.44 | 93.15 | 93.03 | 92.68 |
| | Sentiment Classification | IMDB | 95.97 | 95.27 † | 92.50 | 88.69 | 85.20 | 84.50 |
| Sequence Tagging | Named Entity Recognition | ACE 2005 | 88.75 | 88.75 | 38.51 | 54.78 | 55.27 | 56.03 |
| | | CoNLL 2003 | 92.25 | 91.43 | 51.83 | 59.45 | 61.51 | 74.49 |
| | | OntoNotes v5 | 88.85 | 88.85 | 7.72 | 18.61 | 15.92 | 15.01 |
| | | HONOR | - | - | 45.69 | 56.40 | 57.63 | 64.08 |
| | | MSRANER | - | - | 28.31 | 57.48 | 56.34 | 57.08 |
| | | OntoNote4NER | - | - | 31.92 | 52.04 | 47.99 | 53.74 |
| | Part-of-speech Tagging | WSJ | 97.75 | 97.75 ‡ | 76.80 | 83.83 | 83.56 | 84.60 |
| | | Daily547 | - | - | 65.00 | 83.24 | 83.66 | 84.22 |
| | | PKU-SEGPOS | - | - | 66.94 | 77.29 | 78.16 | 81.13 |
| The Winograd Schema Challenge | The Winograd Schema Challenge | WSC273 | 75.10 | 56.00 | 62.46 | 64.04 | 65.44 | 64.04 |

## B Prompts

For each dataset, we designed three prompts in the 0/1/3/6/9-shot scenario, respectively. Since 3/6/9-shot just adds more in-context compared to 1-shot, we list the prompts we use for each dataset in Table 23 to Table 44 for 0-shot and 1-shot.

Table 23: 0/1-shot prompts for SQuAD1.0 dataset. The "{context}" should be replaced by passage, the "{question}" should be replaced by question.

| # Shot | Prompts |
| --- | --- |
| 0-shot | Passage:{context} // Question: {question} // Referring to the passage above, the correct answer to the given question is |
| | Refer to the passage below and answer the following question: // Passage: {context} // Question: {question} // Answer: |
| | Passage: {context} // Question: {question} // Answer: |
| 1-shot | Passage: 'Architecturally, the school has a Catholic character. Atop the Main Building's gold dome is a golden statue of the Virgin Mary. Immediately in front of the Main Building and facing it, is a copper statue of Christ with arms upraised with the legend "Venite Ad Me Omnes". Next to the Main Building is the Basilica of the Sacred Heart. Immediately behind the basilica is the Grotto, a Marian place of prayer and reflection. It is a replica of the grotto at Lourdes, France where the Virgin Mary reputedly appeared to Saint Bernadette Soubirous in 1858. At the end of the main drive (and in a direct line that connects through 3 statues and the Gold Dome), is a simple, modern stone statue of Mary. ' // Question: 'To whom did the Virgin Mary allegedly appear in 1858 in Lourdes France?' // Referring to the passage above, the correct answer to the given question is // Answer: Saint Bernadette Soubirous // Passage:'{context}' // Question: '{question}' // Referring to the passage above, the correct answer to the given question is |
| | Refer to the passage below and answer the following question: // Passage: 'Architecturally, the school has a Catholic character. Atop the Main Building's gold dome is a golden statue of the Virgin Mary. Immediately in front of the Main Building and facing it, is a copper statue of Christ with arms upraised with the legend "Venite Ad Me Omnes". Next to the Main Building is the Basilica of the Sacred Heart. Immediately behind the basilica is the Grotto, a Marian place of prayer and reflection. It is a replica of the grotto at Lourdes, France where the Virgin Mary reputedly appeared to Saint Bernadette Soubirous in 1858. At the end of the main drive (and in a direct line that connects through 3 statues and the Gold Dome), is a simple, modern stone statue of Mary. ' // Question: 'To whom did the Virgin Mary allegedly appear in 1858 in Lourdes France?' // Answer: Saint Bernadette Soubirous // Refer to the passage below and answer the following question: // Passage: '{context}' // Question: '{question}' // Answer: |
| | Passage: 'Architecturally, the school has a Catholic character. Atop the Main Building's gold dome is a golden statue of the Virgin Mary. Immediately in front of the Main Building and facing it, is a copper statue of Christ with arms upraised with the legend "Venite Ad Me Omnes". Next to the Main Building is the Basilica of the Sacred Heart. Immediately behind the basilica is the Grotto, a Marian place of prayer and reflection. It is a replica of the grotto at Lourdes, France where the Virgin Mary reputedly appeared to Saint Bernadette Soubirous in 1858. At the end of the main drive (and in a direct line that connects through 3 statues and the Gold Dome), is a simple, modern stone statue of Mary. ' // Question: 'To whom did the Virgin Mary allegedly appear in 1858 in Lourdes France?' // Answer: Saint Bernadette Soubirous // Passage: '{context}' // Question: '{question}' // Answer: |

Table 24: 0/1-shot prompts for SQuAD2.0 dataset. The "{context}" should be repaced by passage, and the "{question}" should be replaced by question.

| # Shot | Prompts |
| --- | --- |
| 0-shot | Passage:{context} // Question: {question} // Referring to the passage above, the correct answer to the given question is<br><br>Refer to the passage below and answer the following question: // Passage: {context} // Question: {question} // Answer:<br><br>Passage: {context} // Question: {question} // Answer: |
| 1-shot | Passage: 'Architecturally, the school has a Catholic character. Atop the Main Building's gold dome is a golden statue of the Virgin Mary. Immediately in front of the Main Building and facing it, is a copper statue of Christ with arms upraised with the legend "Venite Ad Me Omnes". Next to the Main Building is the Basilica of the Sacred Heart. Immediately behind the basilica is the Grotto, a Marian place of prayer and reflection. It is a replica of the grotto at Lourdes, France where the Virgin Mary reputedly appeared to Saint Bernadette Soubirous in 1858. At the end of the main drive (and in a direct line that connects through 3 statues and the Gold Dome), is a simple, modern stone statue of Mary. ' // Question: 'To whom did the Virgin Mary allegedly appear in 1858 in Lourdes France?' // Referring to the passage above, the correct answer to the given question is // Answer: Saint Bernadette Soubirous // Passage:'{context}' // Question: '{question}' // Referring to the passage above, the correct answer to the given question is<br><br>Refer to the passage below and answer the following question: // Passage: 'Architecturally, the school has a Catholic character. Atop the Main Building's gold dome is a golden statue of the Virgin Mary. Immediately in front of the Main Building and facing it, is a copper statue of Christ with arms upraised with the legend "Venite Ad Me Omnes". Next to the Main Building is the Basilica of the Sacred Heart. Immediately behind the basilica is the Grotto, a Marian place of prayer and reflection. It is a replica of the grotto at Lourdes, France where the Virgin Mary reputedly appeared to Saint Bernadette Soubirous in 1858. At the end of the main drive (and in a direct line that connects through 3 statues and the Gold Dome), is a simple, modern stone statue of Mary. ' // Question: 'To whom did the Virgin Mary allegedly appear in 1858 in Lourdes France?' // Answer: Saint Bernadette Soubirous // Refer to the passage below and answer the following question: // Passage: '{context}' // Question: '{question}' // Answer:<br><br>Passage: 'Architecturally, the school has a Catholic character. Atop the Main Building's gold dome is a golden statue of the Virgin Mary. Immediately in front of the Main Building and facing it, is a copper statue of Christ with arms upraised with the legend "Venite Ad Me Omnes". Next to the Main Building is the Basilica of the Sacred Heart. Immediately behind the basilica is the Grotto, a Marian place of prayer and reflection. It is a replica of the grotto at Lourdes, France where the Virgin Mary reputedly appeared to Saint Bernadette Soubirous in 1858. At the end of the main drive (and in a direct line that connects through 3 statues and the Gold Dome), is a simple, modern stone statue of Mary. ' // Question: 'To whom did the Virgin Mary allegedly appear in 1858 in Lourdes France?' // Answer: Saint Bernadette Soubirous // Passage: '{context}' // Question: '{question}' // Answer: |

Table 25: 0/1-shot prompts for Tacred dataset. The "{subj}" should be replaced by subject word, the "{obj}" should be replaced by object word, and the "{options}" should be replaced by "person and age, no relation, person and title, organization and top members or employees, organization and country of headquarters, person and parents, person and countries of residence, person and children, organization and alternate names, person and charges, person and cities of residence, person and origin, organization and founded by, person and employee of, person and sibling, person and alternate names, organization and website, person and religion, person and state or province of birth, organization and parents, organization and subsidiaries, person and other family, person and state or provinces of residence, organization and members, person and cause of death, organization and member of, organization and number of employees or members, person and country of birth, organization and shareholders, organization and state or province of headquarters, person and city of death, person and date of birth, person and spouse, organization and city of headquarters, person and date of death, person and schools attended, organization and political or religious affiliation, person and country of death, organization and founded, person and state or province of birth, person and city of birth, organization and dissolved".

| # Shot | Prompts |
| --- | --- |
| 0-shot | '{token}' // In above text, what is the relationship between '{subj}' and '{obj}'? // Options: '{options}' // Answer: |
|  | '{token}' // Determine the relationship between '{subj}' and '{obj}' in above sentence. // Options: '{options}' // Answer: |
|  | '{token}' // Find the relationship between '{subj}' and '{obj}' from above sentence. // Options: '{options}' // Answer: |
| 1-shot | 'Graham , 55 , has maintained his innocence in the killing .' // In above text, what is the relationship between 'Graham' and '55'? // Answer: person and age // '{token}' // In above text, what is the relationship between '{subj}' and '{obj}'? // Options: '{options}' // Answer: |
|  | 'Graham , 55 , has maintained his innocence in the killing .' // Determine the relationship between 'Graham' and '55' in above sentence. // Answer: person and age // '{token}' // Determine the relationship between '{subj}' and '{obj}' in above sentence. // Options: '{options}' // Answer: |
|  | 'Graham , 55 , has maintained his innocence in the killing .' // Find the relationship between 'Graham' and '55' from above sentence. // Answer: person and age // '{token}' // Find the relationship between '{subj}' and '{obj}' from above sentence. // Options: '{options}' // Answer: |

Table 26: 0/1-shot prompts for MNLI-m dataset. The "{premise}" should be replaced by premise, and the "{hypothesis}" should be replaced by hypothesis.

| # Shot | Prompts |
| --- | --- |
| 0-shot | '{premise}' Based on the previous passage, is it entailment or neutral or contradiction that '{hypothesis}' |
| | Suppose '{premise}' Can we infer that '{hypothesis}'? Please choose one answer: entailment, contradiction, neutral |
| | Given that '{premise}' Therefore, it must be entailment or contradiction or neutral that '{hypothesis}' |
| 1-shot | 'He was of two minds, one reveled in the peace of this village.' Based on the previous passage, is it entailment or neutral or contradiction that 'He loved how peaceful the village was.' // Answer: entailment // '{premise}' Based on the previous passage, is it entailment or neutral or contradiction that '{hypothesis}' // Answer: |
| | Suppose 'He was of two minds, one reveled in the peace of this village.' Can we infer that 'He loved how peaceful the village was.'? Please choose one answer: entailment, contradiction, neutral // Answer: entailment // Suppose '{premise}' Can we infer that '{hypothesis}'? Please choose one answer: entailment, contradiction, neutral // Answer: |
| | Given that 'He was of two minds, one reveled in the peace of this village.' Therefore, it must be entailment or contradiction or neutral that 'He loved how peaceful the village was.' // Answer: entailment // Given that '{premise}' Therefore, it must be entailment or contradiction or neutral that '{hypothesis}' // Answer: |

Table 27: 0/1-shot prompts for MNLI-mm dataset. The "{premise}" should be replaced by the premise, and the "{hypothesis}" should be replaced by the hypothesis.

| # Shot | Prompts |
| --- | --- |
| 0-shot | '{premise}' Based on the previous passage, is it entailment or neutral or contradiction that '{hypothesis}' <br><br> Suppose '{premise}' Can we infer that '{hypothesis}'? Please choose one answer: entailment, contradiction, neutral <br><br> Given that '{premise}' Therefore, it must be entailment or contradiction or neutral that '{hypothesis}' |
| 1-shot | 'I'll twist him, sir.' Based on the previous passage, is it entailment or neutral or contradiction that 'I'll make him straight.' // Answer: contradiction // '{premise}' Based on the previous passage, is it entailment or neutral or contradiction that '{hypothesis}' // Answer: <br><br> Suppose 'I'll twist him, sir.' Can we infer that 'I'll make him straight.'? Please choose one answer: entailment, contradiction, neutral // Answer: contradiction // Suppose '{premise}' Can we infer that '{hypothesis}'? Please choose one answer: entailment, contradiction, neutral // Answer: <br><br> Given that 'I'll twist him, sir.' Therefore, it must be entailment or contradiction or neutral that 'I'll make him straight.' // Answer: contradiction // Given that '{premise}' Therefore, it must be entailment or contradiction or neutral that '{hypothesis}' // Answer:' |

Table 28: 0/1-shot prompts for SNLI dataset. The "{premise}" should be replaced by the premise, and the "{hypothesis}" should be replaced by the hypothesis.

| # Shot | Prompts |
| --- | --- |
| 0-shot | '{premise}' Based on the previous passage, is it entailment or neutral or contradiction that '{hypothesis}' |
| | Suppose '{premise}' Can we infer that '{hypothesis}'? Please choose one answer: entailment, contradiction, neutral |
| | Given that '{premise}' Therefore, it must be entailment or contradiction or neutral that '{hypothesis}' |
| 1-shot | Premise:'A person on a horse jumps over a broken down airplane.'. Based on this premise, can we conclude the hypothesis 'A person is training his horse for a competition.' is true? // Options: neutral, contradiction, entailment // Answer: neutral // Premise:'{premise}'. Based on this premise, can we conclude the hypothesis '{hypothesis}' is true? // Options: neutral, contradiction, entailment // Answer: |
| | Suppose 'A person on a horse jumps over a broken down airplane.' // Can we infer that 'A person is training his horse for a competition.'? // Options: neutral, contradiction, entailment // Answer: neutral // Suppose '{premise}' // Can we infer that '{hypothesis}'? // options: neutral,contradiction,entailment // Answer: |
| | Given that 'A person on a horse jumps over a broken down airplane.' Therefore, it must be true that 'A person is training his horse for a competition.'? // Options: neutral, contradiction, entailment // Answer: neutral // Given that '{premise}' Therefore, it must be true that '{hypothesis}'? // options:neutral,contradiction,entailment // Answer: |

Table 29: 0/1-shot prompts for MRPC dataset. The "{sentence1}" should be replaced by the first sentence, and the "{sentence2}" should be replaced by the second sentence.

| # Shot | Prompts |
| --- | --- |
| 0-shot | Does the sentence '{sentence1}' paraphrase (that is, mean the same thing as) this sentence? '{sentence2}' // Options: yes, no<br><br>I want to know whether the following two sentences mean the same thing. '{sentence1}' '{sentence2}' // Options: yes, no<br><br>Do the following two sentences mean the same thing? // '{sentence1}' // '{sentence2}' // Options: yes, no |
| 1-shot | Does the sentence 'Amrozi accused his brother , whom he called " the witness " , of deliberately distorting his evidence .' paraphrase (that is, mean the same thing as) this sentence? 'Referring to him as only " the witness " , Amrozi accused his brother of deliberately distorting his evidence .' // Options: yes, no // Answer: Yes // Does the sentence '{sentence1}' paraphrase (that is, mean the same thing as) this sentence? '{sentence2}' // Options: yes, no // Answer:<br><br>I want to know whether the following two sentences mean the same thing. 'Amrozi accused his brother , whom he called " the witness " , of deliberately distorting his evidence .' 'Referring to him as only " the witness " , Amrozi accused his brother of deliberately distorting his evidence .' // Options: yes, no // Answer: Yes // I want to know whether the following two sentences mean the same thing. '{sentence1}' '{sentence2}' // Options: yes, no // Answer:<br><br>Do the following two sentences mean the same thing? // 'Amrozi accused his brother , whom he called " the witness " , of deliberately distorting his evidence .' // 'Referring to him as only " the witness " , Amrozi accused his brother of deliberately distorting his evidence .' // Options: yes, no // Answer: Yes // Do the following two sentences mean the same thing? // '{sentence1}' // '{sentence2}' // Options: yes, no // Answer: |

Table 30: 0/1-shot prompts for QQP dataset. The "{question1}" should be replaced by the first question, and the "{question}" should be replaced by the second question.

| # Shot | Prompts |
| --- | --- |
| 0-shot | Can an answer to '{question1}' also be used to answer '{question2}'?<br><br>Are the questions '{question1}' and '{question2}' asking the same thing? Yes or no?<br><br>I want to know whether the following two questions mean the same thing. '{question1}' '{question2}' Do they? // Options: yes, no |
| 1-shot | Can an answer to 'What is the step by step guide to invest in share market in india?' also be used to answer 'What is the step by step guide to invest in share market?'? // Answer: No // Can an answer to '{question1}' also be used to answer '{question2}'? // Answer:<br><br>Are the questions 'What is the step by step guide to invest in share market in india?' and 'What is the step by step guide to invest in share market?' asking the same thing? Yes or no? // Answer: No // Are the questions '{question1}' and '{question2}' asking the same thing? Yes or no? // Answer:<br><br>I want to know whether the following two questions mean the same thing. 'What is the step by step guide to invest in share market in india?' 'What is the step by step guide to invest in share market?' Do they? // Options: yes, no // Answer: No // I want to know whether the following two questions mean the same thing. '{question1}' '{question2}' Do they? // Options: yes, no // Answer: |

Table 31: 0/1-shot prompts for SemEval2014-Laptop dataset. The "{aspect}" should be replaced by the aspect to be analyzed, and the "{sentence}" should be replaced by a sentence.

| # Shot | Prompts |
| --- | --- |
| 0-shot | Analyze the sentiment towards the '{aspect}' of '{sentence}' and determine if it is positive, negative, or neutral. |
| | What is the sentiment towards '{sentence}' in terms of '{aspect}'? Are they viewed positively, negatively, or neutrally? |
| | '{sentence}' Express your sentiment towards the aspect of '{aspect}' using positive, negative, or neutral. |
| 1-shot | Analyze the sentiment towards the 'BIOS' of 'But sadly the replacement froze-up while updating the BIOS again and shut down and would not turn back on.' and determine if it is positive, negative, or neutral. // Answer: negative // Analyze the sentiment towards the '{aspect}' of '{sentence}' and determine if it is positive, negative, or neutral. // Answer: |
| | What is the sentiment towards 'But sadly the replacement froze-up while updating the BIOS again and shut down and would not turn back on.' in terms of 'BIOS'? Are they viewed positively, negatively, or neutrally? // Answer: negative // What is the sentiment towards '{sentence}' in terms of '{aspect}'? Are they viewed positively, negatively, or neutrally? // Answer: |
| | 'But sadly the replacement froze-up while updating the BIOS again and shut down and would not turn back on.' Express your sentiment towards the aspect of 'BIOS' using positive, negative, or neutral. // Answer: negative // '{sentence}' Express your sentiment towards the aspect of '{aspect}' using positive, negative, or neutral. // Answer: |

Table 32: 0/1-shot prompts for SemEval2014-Restaurant dataset. The "{aspect}" should be replaced by the aspect to be analyzed, and the "{sentence}" should be replaced by the sentence.

| # Shot | Prompts |
| --- | --- |
| 0-shot | Analyze the sentiment towards the '{aspect}' of '{sentence}' and determine if it is positive, negative, or neutral.<br><br>What is the sentiment towards '{sentence}' in terms of '{aspect}'? Are they viewed positively, negatively, or neutrally?<br><br>'{sentence}' Express your sentiment towards the aspect of '{aspect}' using positive, negative, or neutral. |
| 1-shot | Analyze the sentiment towards the 'dishes' of 'The food is good, especially their more basic dishes, and the drinks are delicious.' and determine if it is positive, negative, or neutral. // Answer: positive // Analyze the sentiment towards the '{aspect}' of '{sentence}' and determine if it is positive, negative, or neutral. // Answer:<br><br>What is the sentiment towards 'The food is good, especially their more basic dishes, and the drinks are delicious.' in terms of 'dishes'? Are they viewed positively, negatively, or neutrally? // Answer: positive // What is the sentiment towards '{sentence}' in terms of '{aspect}'? Are they viewed positively, negatively, or neutrally? // Answer:<br><br>'The food is good, especially their more basic dishes, and the drinks are delicious.' Express your sentiment towards the aspect of 'dishes' using positive, negative, or neutral. // Answer: positive // '{sentence}' Express your sentiment towards the aspect of '{aspect}' using positive, negative, or neutral. // Answer: |

Table 33: 0/1-shot prompts for IMDB dataset. The "{sentence}" should be replaced by sentence.

| # Shot | Prompts |
| --- | --- |
| 0-shot | The sentiment expressed for the movie is positive or negative? // '{sentence}' // options:positive,negative<br><br>The following movie review expresses what sentiment? positive or negative? // '{sentence}' // options:positive,negative<br><br>What is the sentiment expressed by the reviewer for the movie? positive or negative? // '{sentence}' // options:positive,negative |
| 1-shot | The sentiment expressed for the movie is positive or negative? // 'The Great Dictator is a beyond-excellent film. Charlie Chaplin succeeds in being both extremely funny and witty and yet at the same time provides a strong statement in his satire against fascism. The anti-Nazi speech by Chaplin at the end, with its values, is one of filmdom's great moments. Throughout this movie, I sensed there was some higher form of intelligence, beyond genuinely intelligent filmmaking, at work.' // options:positive,negative // Answer: positive // '{sentence}' // options:positive,negative // Answer:<br><br>The following movie review expresses what sentiment? positive or negative? // 'The Great Dictator is a beyond-excellent film. Charlie Chaplin succeeds in being both extremely funny and witty and yet at the same time provides a strong statement in his satire against fascism. The anti-Nazi speech by Chaplin at the end, with its values, is one of filmdom's great moments. Throughout this movie, I sensed there was some higher form of intelligence, beyond genuinely intelligent filmmaking, at work.' // Answer: positive // '{sentence}' // options:positive,negative // Answer:<br><br>What is the sentiment expressed by the reviewer for the movie? positive or negative? // 'The Great Dictator is a beyond-excellent film. Charlie Chaplin succeeds in being both extremely funny and witty and yet at the same time provides a strong statement in his satire against fascism. The anti-Nazi speech by Chaplin at the end, with its values, is one of filmdom's great moments. Throughout this movie, I sensed there was some higher form of intelligence, beyond genuinely intelligent filmmaking, at work.' // Answer: positive // '{sentence}' // options:positive,negative // Answer: |

Table 34: 0/1-shot prompts for ACE 2005 dataset. The "{text}" should be replaced by sentence.

| # Shot | Prompts |
| --- | --- |
| 0-shot | Please identify Organization, Person, Geo-political Entity, Facility, Location, Vehicle and Weapon Entity from the given text, output using the format as "Entity: Organization: None\|Person: Word1\|Geo-political Entity: None\|Facility: Word2\|Location: Word3, Word4\|Vehicle: None\|Weapon: None" // Text: {text} // Entity:<br><br>Please list all Organization, Person, Geo-political Entity, Facility, Location, Vehicle and Weapon Entity in the given text, output using the format as "Entity: Organization: None\|Person: Word1\|Geo-political Entity: None\|Facility: Word2\|Location: Word3, Word4\|Vehicle: None\|Weapon: None" // Text: {text} // Entity:<br><br>Extract all Organization, Person, Geo-political Entity, Facility, Location, Vehicle and Weapon Entity from the given text, output using the format as "Entity: Organization: None\|Person: Word1\|Geo-political Entity: None\|Facility: Word2\|Location: Word3, Word4\|Vehicle: None\|Weapon: None" // Text: {text} // Entity: |
| 1-shot | Please identify Organization, Person, Geo-political Entity, Facility, Location, Vehicle and Weapon Entity from the given text // Text: thank you , paula . // Entity: Organization: None\|Person: you, paula\|Geo-political Entity: None\|Facility: None\|Location: None\|Vehicle: None\|Weapon: None // Text: {text} // Entity:<br><br>Please list all Organization, Person, Geo-political Entity, Facility, Location, Vehicle and Weapon Entity in the given text // Text: thank you , paula . // Entity: Organization: None\|Person: you, paula\|Geo-political Entity: None\|Facility: None\|Location: None\|Vehicle: None\|Weapon: None // Text: {text} // Entity:<br><br>Extract all Organization, Person, Geo-political Entity, Facility, Location, Vehicle and Weapon Entity from the given text // Text: thank you , paula . // Entity: Organization: None\|Person: you, paula\|Geo-political Entity: None\|Facility: None\|Location: None\|Vehicle: None\|Weapon: None // Text: {text} // Entity: |

Table 35: 0/1-shot prompts for CoNLL 2003 dataset. The "{text}" should be replaced by text.

| # Shot | Prompts |
| --- | --- |
| 0-shot | Please identify Organization, Person, Location and Miscellaneous Entity from the given text, output using the format as "Entity: Organization: None\|Person: None\|Location: Word1, Word2\|Miscellaneous: Word3" // Text: {text} // Entity:<br><br>Please list all Organization, Person, Location and Miscellaneous Entity in the given text, output using the format as "Entity: Organization: None\|Person: None\|Location: Word1, Word2\|Miscellaneous: Word3" // Text: {text} // Entity:<br><br>Extract all Organization, Person, Location and Miscellaneous Entity from the given text, output using the format as "Entity: Organization: None\|Person: None\|Location: Word1, Word2\|Miscellaneous: Word3" // Text: {text} // Entity: |
| 1-shot | Please identify Organization, Person, Location and Miscellaneous Entity from the given text // Text: AL-AIN , United Arab Emirates 1996-12-06 // Entity: Organization: None\|Person: None\|Location: AL-AIN, United Arab Emirates\|Miscellaneous: None // Text: {text} // Entity:<br><br>Please list all Organization, Person, Location and Miscellaneous Entity in the given text // Text: AL-AIN , United Arab Emirates 1996-12-06 // Entity: Organization: None\|Person: None\|Location: AL-AIN, United Arab Emirates\|Miscellaneous: None // Text: {text} // Entity:<br><br>Extract all Organization, Person, Location and Miscellaneous Entity from the given text // Text: AL-AIN , United Arab Emirates 1996-12-06 // Entity: Organization: None\|Person: None\|Location: AL-AIN, United Arab Emirates\|Miscellaneous: None // Text: {text} // Entity: |

Table 36: 0/1-shot prompts for OntoNotes v5 dataset. The "{text}" should be replaced by sentence and the "{format}" should be replaced by "Entity: Organization: None|Person: Word1|Geo-political Entity: None|Facility: None|Location: Word2|Time: Word3|Cardinal: None|Money: None|Date: None|Percent: None|Language: None|Work of art: None|Nationalities or religious or political groups: Word4, Word5|Quantity: None|Ordinal: None|Product: None|Event: None|Law: None".

| # Shot | Prompts |
| --- | --- |
| 0-shot | Please identify Organization, Person, Geo-political Entity, Facility, Location, Time, Cardinal, Money, Date, Percent, Language, Work of art, Nationalities or religious or political groups, Quantity, Ordinal, Product, Event, Law Entity from the given text, output using the format as '{format}' // Text: {text} // Entity: |
| | Please list all Organization, Person, Geo-political Entity, Facility, Location, Time, Cardinal, Money, Date, Percent, Language, Work of art, Nationalities or religious or political groups, Quantity, Ordinal, Product, Event, Law Entity in the given text, output using the format as '{format}' // Text: {text} // Entity: |
| | Extract all Organization, Person, Geo-political Entity, Facility, Location, Time, Cardinal, Money, Date, Percent, Language, Work of art, Nationalities or religious or political groups, Quantity, Ordinal, Product, Event, Law Entity from the given text, output using the format as '{format}' // Text: {text} // Entity: |
| 1-shot | Please identify Organization, Person, Geo-political Entity, Facility, Location, Time, Cardinal, Money, Date, Percent, Language, Work of art, Nationalities or religious or political groups, Quantity, Ordinal, Product, Event, Law Entity from the given text // Text: Graphic by Tsai Chih - pen // Entity: Organization: None|Person: Tsai Chih - pen|Geo-political Entity: None|Facility: None|Location: None|Time: None|Cardinal: None|Money: None|Date: None|Percent: None|Language: None|Work of art: None|Nationalities or religious or political groups: None|Quantity: None|Ordinal: None|Product: None|Event: None|Law: None // Text: {text} // Entity: |
| | Please list all Organization, Person, Geo-political Entity, Facility, Location, Time, Cardinal, Money, Date, Percent, Language, Work of art, Nationalities or religious or political groups, Quantity, Ordinal, Product, Event, Law Entity in the given text // Text: Graphic by Tsai Chih - pen // Entity: Organization: None|Person: Tsai Chih - pen|Geo-political Entity: None|Facility: None|Location: None|Time: None|Cardinal: None|Money: None|Date: None|Percent: None|Language: None|Work of art: None|Nationalities or religious or political groups: None|Quantity: None|Ordinal: None|Product: None|Event: None|Law: None // Text: {text} // Entity: |
| | Extract all Organization, Person, Geo-political Entity, Facility, Location, Time, Cardinal, Money, Date, Percent, Language, Work of art, Nationalities or religious or political groups, Quantity, Ordinal, Product, Event, Law Entity from the given text // Text: Graphic by Tsai Chih - pen // Entity: Organization: None|Person: Tsai Chih - pen|Geo-political Entity: None|Facility: None|Location: None|Time: None|Cardinal: None|Money: None|Date: None|Percent: None|Language: None|Work of art: None|Nationalities or religious or political groups: None|Quantity: None|Ordinal: None|Product: None|Event: None|Law: None // Text: {text} // Entity: |

Table 37: 0/1-shot prompts for HONOR dataset. The "{text}" should be replaced by sentence.

| # Shot | Prompts |
| --- | --- |
| 0-shot | 请识别文本中的所有结束日期、参与人、开始日期、开始时间、开始结束时间、发生地、结束时间、开始结束日期，每个词最多出现在一个类别，使用"结果：结束日期：无，参与人：无，开始日期：词语1，开始时间：词语2，开始结束时间：无，发生地：无，结束时间：词语4，开始结束日期：无"的格式输出 // 文本：{text} // 结果： <br><br> 请从文本中识别结束日期、参与人、开始日期、开始时间、开始结束时间、发生地、结束时间、开始结束日期并列举出来，每个词最多出现在一个类别，使用"结果：结束日期：无，参与人：无，开始日期：词语1，开始时间：词语2，开始结束时间：无，发生地：无，结束时间：词语4，开始结束日期：无"的格式输出 // 文本：{text} // 结果： <br><br> 请告诉我给定文本中的结束日期、参与人、开始日期、开始时间、开始结束时间、发生地、结束时间、开始结束日期是什么，每个词最多出现在一个类别，使用"结果：结束日期：无，参与人：无，开始日期：词语1，开始时间：词语2，开始结束时间：无，发生地：无，结束时间：词语4，开始结束日期：无"的格式输出 // 文本：{text} // 结果： |
| 1-shot | 请识别文本中的所有结束日期、参与人、开始日期、开始时间、开始结束时间、发生地、结束时间、开始结束日期，每个词最多出现在一个类别 // 文本：让他帮我参加一下每天早上的会 // 结果：结束日期：无，参与人：无，开始日期：每天，开始时间：早上，开始结束时间：无，发生地：无，结束时间：无，开始结束日期：无 // 文本：{text} // 结果： <br><br> 请从文本中识别结束日期、参与人、开始日期、开始时间、开始结束时间、发生地、结束时间、开始结束日期并列举出来，每个词最多出现在一个类别 // 文本：让他帮我参加一下每天早上的会 // 结果：结束日期：无；参与人：无；开始日期：每天；开始时间：早上；开始结束时间：无；发生地：无；结束时间：无；开始结束日期：无 // 文本：{text} // 结果： <br><br> 请告诉我给定文本中的结束日期、参与人、开始日期、开始时间、开始结束时间、发生地、结束时间、开始结束日期是什么，每个词最多出现在一个类别 // 文本：让他帮我参加一下每天早上的会 // 结果：结束日期：无；参与人：无；开始日期：每天；开始时间：早上；开始结束时间：无；发生地：无；结束时间：无；开始结束日期：无 // 文本：{text} // 结果： |

Table 38: 0/1-shot prompts for MSRANER dataset. The "{text}" should be replaced by sentence.

| # Shot | Prompts |
| --- | --- |
| 0-shot | 请识别文本中的所有人名、地名、组织名，每个词最多出现在一个类别，使用"结果：人名：词语1，词语2; 地名：无; 组织名：词语3"的格式输出 // 文本：{text} // 结果： |
| | 请从给定文本中识别人名、地名、组织名并列举出来，每个词最多出现在一个类别，使用"结果：人名：词语1，词语2; 地名：无; 组织名：词语3"的格式输出 // 文本：{text} // 结果： |
| | 请告诉我给定文本中的人名、地名、组织名是什么，每个词最多出现在一个类别，使用"结果：人名：词语1，词语2; 地名：无; 组织名：词语3"的格式输出 // 文本：{text} // 结果： |
| 1-shot | 请识别文本中的所有人名、地名、组织名，每个词最多出现在一个类别 // 文本：中共中央致中国致公党十一大的贺词 // 结果：人名：无; 地名：无; 组织名：中共中央，中国致公党十一大 // 文本：{text} // 结果： |
| | 请从给定文本中识别人名、地名、组织名并列举出来，每个词最多出现在一个类别 // 文本：中共中央致中国致公党十一大的贺词 // 结果：人名：无; 地名：无; 组织名：中共中央，中国致公党十一大 // 文本：{text} // 结果： |
| | 请告诉我给定文本中的人名、地名、组织名是什么，每个词最多出现在一个类别 // 文本：中共中央致中国致公党十一大的贺词 // 结果：人名：无; 地名：无; 组织名：中共中央，中国致公党十一大 // 文本：{text} // 结果： |

Table 39: 0/1-shot prompts for OntoNote4NER dataset. $ is used as a separator. The "{text}" should be replaced by sentence.

| # Shot | Prompts |
| --- | --- |
| 0-shot | 请识别文本中的所有地缘政治实体、地名、组织机构名、人名，每个词最多出现在一个类别，使用"结果：地缘政治实体 $$ 词语 1$ 词语 2$ 词语 3$ 词语 4$$$ 地名 $$ 无 $$$ 组织名 $$ 无 $$$ 人名 $$ 词语 5"的格式输出 // 文本：{text} // 结果 <br><br> 请从文本中识别地缘政治实体、地名、组织机构名、人名并列举出来，每个词最多出现在一个类别，使用"结果：地缘政治实体 $$ 词语 1$ 词语 2$ 词语 3$ 词语 4$$$ 地名 $$ 无 $$$ 组织名 $$ 无 $$$ 人名 $$ 词语 5"的格式输出 // 文本：{text} // 结果 <br><br> 请告诉我给定文本中的地缘政治实体、地名、组织机构名、人名是什么，每个词最多出现在一个类别，每个词最多出现在一个类别，使用"结果：地缘政治实体 $$ 词语 1$ 词语 2$ 词语 3$ 词语 4$$$ 地名 $$ 无 $$$ 组织名 $$ 无 $$$ 人名 $$ 词语 5"的格式输出 // 文本：{text} // 结果 |
| 1-shot | 请识别文本中的所有地缘政治实体、地名、组织机构名、人名，每个词最多出现在一个类别 // 文本：二次大战日本结束统治后，台湾回归中国，帛琉则成为美国的托管地，并于一九八〇年代开始与我国有了政治上的接触。// 结果：地缘政治实体 $$ 日本 $ 台湾 $ 中国 $ 帛琉 $$$ 地名 $$ 无 $$$ 组织名 $$ 无 $$$ 人名 $$ 无 // 文本：{text} // 结果 <br><br> 请从文本中识别地缘政治实体、地名、组织机构名、人名并列举出来，每个词最多出现在一个类别 // 文本：二次大战日本结束统治后，台湾回归中国，帛琉则成为美国的托管地，并于一九八〇年代开始与我国有了政治上的接触。// 结果：地缘政治实体 $$ 日本 $ 台湾 $ 中国 $ 帛琉 $$$ 地名 $$ 无 $$$ 组织名 $$ 无 $$$ 人名 $$ 无 // 文本：{text} // 结果： <br><br> 请告诉我给定文本中的地缘政治实体、地名、组织机构名、人名是什么，每个词最多出现在一个类别 // 文本：二次大战日本结束统治后，台湾回归中国，帛琉则成为美国的托管地，并于一九八〇年代开始与我国有了政治上的接触。// 结果：地缘政治实体 $$ 日本 $ 台湾 $ 中国 $ 帛琉 $$$ 地名 $$ $$$ 组织名 $$ 无 $$$ 人名 $$ 无 // 文本：{text} // 结果： |

Table 40: 0/1-shot prompts for WSJ dataset. ¥ is used as a separator. The "{text}" should be replaced by sentence and the "{candidate}" should be replaced by "adjective, plural noun, preposition or conjunction, determiner, singular proper noun, coordinating conjunction, past tense verb, singular or mass noun, wh-determiner, modal, base form verb, wh-adverb, comma, gerund or present partical, to, possessive ending, sentence boundary marker, possessive wh-pronoun, non-3rd person singular present verb, left round bracket, right round bracket, adverb, past participle verb, 3rd person singular present verb, left double quotation mark, right double quotation mark, comparative adverb, monetary values, cardinal number, comparative adjective, particle, personal pronoun, colon character, possessive pronoun, predeterminer, superlative adverb, wh-pronoun, superlative adjective, foreign word, list marker, interjection, existential there, pound symbol, plural proper noun, symbol".

| # Shot | Prompts |
| --- | --- |
| 0-shot | Do part-of-speech task for the given text using the categories in candidate list, output using the format as "Word1¥Categary¥¥Word2¥Categary¥¥Word3¥Category" // Candidate list: {candidate} // Text: {text} // Result:<br><br>Tag the parts of speech in the given text using the categories in candidate list, output using the format as "Word1¥Categary¥¥Word2¥Categary¥¥Word3¥Category" // Candidate list: {candidate} // Text: {text} // Result:<br><br>Label the words in the given text using the categories in candidate list, output using the format as "Word1¥Categary¥¥Word2¥Categary¥¥Word3¥Category" // Candidate list: {candidate} // Text: {text} // Result: |
| 1-shot | Do part-of-speech task for the given text using the categories in candidate list // Candidate list: {candidate} // Text: Few changes were made in the way the markets are regulated . // Result: Few¥adjective¥¥changes¥plural noun¥¥were¥past tense verb¥¥made¥past participle verb¥¥in¥preposition or conjuction¥¥the¥determiner¥¥way¥singular or mass noun¥¥the¥determiner¥¥markets¥plural noun¥¥are¥non-3rd person singular present verb¥¥regulated¥past participle verb¥¥.¥sentence boundary marker // Text: {text} // Result:<br><br>Tag the parts of speech in the given text using the categories in candidate list // Candidate list: {candidate} // Text: Few changes were made in the way the markets are regulated . // Result: Few¥adjective¥¥changes¥plural noun¥¥were¥past tense verb¥¥made¥past participle verb¥¥in¥preposition or conjuction¥¥the¥determiner¥¥way¥singular or mass noun¥¥the¥determiner¥¥markets¥plural noun¥¥are¥non-3rd person singular present verb¥¥regulated¥past participle verb¥¥.¥sentence boundary marker // Text: {text} // Result:<br><br>Label the words in the given text using the categories in candidate list // Candidate list: {candidate} // Text: Few changes were made in the way the markets are regulated . // Result: Few¥adjective¥¥changes¥plural noun¥¥were¥past tense verb¥¥made¥past participle verb¥¥in¥preposition or conjuction¥¥the¥determiner¥¥way¥singular or mass noun¥¥the¥determiner¥¥markets¥plural noun¥¥are¥non-3rd person singular present verb¥¥regulated¥past participle verb¥¥.¥sentence boundary marker // Text: {text} // Result: |

Table 41: 1-shot prompts for WSJ dataset using original lables. The "{text}" should be replaced by sentence.

| # Shot | Prompts |
| --- | --- |
| 1-shot | Do part-of-speech task for the given X from WSJ dataset // X: " That 's baseball . // Y: """_""" "That"_"DT" "'s"_"VBZ" "baseball"_"NN" "."_"." // X: {text} // Y: |
| | Annotation for the given X // X: " That 's baseball . // Y: """_""" "That"_"DT" "'s"_"VBZ" "baseball"_"NN" "."_"." // X: {text} // Y: |
| | Grammatical tagging of X // X: " That 's baseball . // Y: """_""" "That"_"DT" "'s"_"VBZ" "baseball"_"NN" "."_"." // X: {text} // Y: |

Table 42: 0/1-shot prompts for Daily547 dataset. ¥ is used as a separator. The "{text}" should be replaced by sentence and the "{candidate}" should be replaced by "common noun, pronoun, proper noun, nominal + possessive, proper noun + possessive, verb incl. copula and auxiliaries, adjective, adverb, interjection, determine, pre- or postposition/subordinating conjunction, coordinating conjunction, verb partical, existential there/predeterminers, hashtag, at-mention, discourse marker, URL/email address, emoticon, numeral, punctuation, other, nominal + verbal, proper noun + verbal, X + verbal".

| # Shot | Prompts |
| --- | --- |
| 0-shot | Do part-of-speech task for the given text using the categories in candidate list, output using the format as "Word1¥Category¥¥Word2¥Category¥¥Word3¥Category" // Candidate list: {candidate} // Text: {text} // Result: <br><br> Tag the parts of speech in the given text using the categories in candidate list, output using the format as "Word1¥Category¥¥Word2¥Category¥¥Word3¥Category" // Candidate list: {candidate} // Text: {text} // Result: <br><br> Label the words in the given text using the categories in candidate list, output using the format as "Word1¥Category¥¥Word2¥Category¥¥Word3¥Category" // Candidate list: {candidate} // Text: {text} // Result: |
| 1-shot | Do part-of-speech task for the given text using the categories in candidate list // Candidate list: {candidate} // Text: Bridalplasty ! Love this showww . // Result: Bridalplasty¥proper noun¥¥!¥punctuation¥¥Love¥verb incl. copula and auxiliaries¥¥this¥determine¥¥showww¥common noun¥¥.¥punctuation // Text: {text} // Result: <br><br> Tag the parts of speech in the given text using the categories in candidate list // Candidate list: {candidate} // Text: Bridalplasty ! Love this showww . // Result: Bridalplasty¥proper noun¥¥!¥punctuation¥¥Love¥verb incl. copula and auxiliaries¥¥this¥determine¥¥showww¥common noun¥¥.¥punctuation // Text: {text} // Result: <br><br> Label the words in the given text using the categories in candidate list // Candidate list: {candidate} // Text: Bridalplasty ! Love this showww . // Result: Bridalplasty¥proper noun¥¥!¥punctuation¥¥Love¥verb incl. copula and auxiliaries¥¥this¥determine¥¥showww¥common noun¥¥.¥punctuation // Text: {text} // Result: |

Table 43: 0/1-shot prompts for PKU-SEGPOS dataset. The "{text}" should be replaced by sentence.

| # Shot | Prompts |
| --- | --- |
| 0-shot | 请使用候选集的词性，对给定文本中的每个词语进行标注, 使用" 词语 1_ 词性/词语 2_ 词性/词语 3_ 词性" 的格式输出 // 候选集：名词, 时间词, 处所词, 方位词, 数词, 量词, 区别词, 代词, 动词, 形容词, 状态词, 副词, 介词, 连词, 助词, 语气词, 叹词, 拟声词, 成语, 习用语, 简称词, 前接成分, 后接成分, 语素, 非语素字, 标点符号 // 文本：{text} // 结果： <br><br> 请使用候选集标注给定文本中每个词的词性, 使用" 词语 1_ 词性/词语 2_ 词性/词语 3_ 词性" 的格式输出 // 候选集：名词, 时间词, 处所词, 方位词, 数词, 量词, 区别词, 代词, 动词, 形容词, 状态词, 副词, 介词, 连词, 助词, 语气词, 叹词, 拟声词, 成语, 习用语, 简称词, 前接成分, 后接成分, 语素, 非语素字, 标点符号 // 文本：{text} // 结果： <br><br> 对于给定的文本，请使用候选集标注每个词的词性, 使用" 词语 1_ 词性/词语 2_ 词性/词语 3_ 词性" 的格式输出 // 候选集：名词, 时间词, 处所词, 方位词, 数词, 量词, 区别词, 代词, 动词, 形容词, 状态词, 副词, 介词, 连词, 助词, 语气词, 叹词, 拟声词, 成语, 习用语, 简称词, 前接成分, 后接成分, 语素, 非语素字, 标点符号 // 文本：{text} // 结果： |
| 1-shot | 请使用候选集的词性，对给定文本中的每个词语进行标注 // 候选集：名词, 时间词, 处所词, 方位词, 数词, 量词, 区别词, 代词, 动词, 形容词, 状态词, 副词, 介词, 连词, 助词, 语气词, 叹词, 拟声词, 成语, 习用语, 简称词, 前接成分, 后接成分, 语素, 非语素字, 标点符号 // 文本：天津开发区蒸蒸日上。// 结果：天津 _ 名词/开发区 _ 名词/蒸蒸日上 _ 成语/。_ 标点符号 // 文本：{text} // 结果： <br><br> 请使用候选集标注给定文本中每个词的词性 // 候选集：名词, 时间词, 处所词, 方位词, 数词, 量词, 区别词, 代词, 动词, 形容词, 状态词, 副词, 介词, 连词, 助词, 语气词, 叹词, 拟声词, 成语, 习用语, 简称词, 前接成分, 后接成分, 语素, 非语素字, 标点符号 // 文本：天津开发区蒸蒸日上。// 结果：天津 _ 名词/开发区 _ 名词/蒸蒸日上 _ 成语/。_ 标点符号 // 文本：{text} // 结果： <br><br> 对于给定的文本，请使用候选集标注每个词的词性 // 候选集：名词, 时间词, 处所词, 方位词, 数词, 量词, 区别词, 代词, 动词, 形容词, 状态词, 副词, 介词, 连词, 助词, 语气词, 叹词, 拟声词, 成语, 习用语, 简称词, 前接成分, 后接成分, 语素, 非语素字, 标点符号 // 文本：天津开发区蒸蒸日上。// 结果：天津 _ 名词/开发区 _ 名词/蒸蒸日上 _ 成语/。_ 标点符号 // 文本：{text} // 结果： |

Table 44: 0/1-shot prompts for WSC273 dataset. The "{text}" should be replaced by sentence, the "{target 1}" should be replaced by the first target, and the "{target 2}" should be replaced by the second target.

| # Shot | Prompts |
| --- | --- |
| 0-shot | '{text}' // In the previous sentences, does '{target 2}' refer to '{target 1}'? // Options: yes, no // Answer: |
|  | '{text}' // Here, does '{target 2}' stand for '{target 1}'? // Options: yes, no // Answer: |
|  | '{text}' // In the passage above, can '{target 2}' be replaced by '{target 1}'? // Options: yes, no // Answer: |
| 1-shot | 'the board of aldermen refused the protesters a permit because they feared violence .' // In the previous sentences, does 'they' refer to 'the board of aldermen'? // Answer: yes // '{text}' // In the previous sentences, does '{target 2}' refer to '{target 1}'? // Options: yes, no // Answer: |
|  | 'the board of aldermen refused the protesters a permit because they feared violence .' // Here, does 'they' stand for 'the board of aldermen'? // Answer: yes // '{text}' // Here, does '{target 2}' stand for '{target 1}'? // Options: yes, no // Answer: |
|  | 'the board of aldermen refused the protesters a permit because they feared violence .' // In the passage above, can 'they' be replaced by 'the board of aldermen'? // Answer: yes // '{text}' // In the passage above, can '{target 2}' be replaced by '{target 1}'? // Options: yes, no // Answer: |